\documentclass[10pt,twocolumn,letterpaper]{article}

\usepackage{iccv}              % To produce the CAMERA-READY version
\usepackage{times}
\usepackage{epsfig}
\usepackage{graphicx}
\usepackage{amsmath}
\usepackage{amssymb}
\usepackage{comment}

\usepackage{times}
\usepackage{epsfig}
\usepackage{graphicx}
\usepackage{amsmath}
\usepackage{amssymb}
\usepackage[numbers,sort&compress]{natbib}
\usepackage{color}
\usepackage{multirow}
\usepackage{xcolor}

\usepackage{soul}
\usepackage{colortbl}
\usepackage{bbding}
\usepackage{booktabs, colortbl}
\usepackage{longtable}
\usepackage{hhline}
\usepackage[toc]{multitoc}

\usepackage{algorithm}
\usepackage{algorithmic}

\usepackage{todonotes} %
\usepackage{enumitem}
\usepackage{bbm}

\usepackage{minitoc}
\usepackage[toc,page,header]{appendix}
\definecolor{myblue}{HTML}{0072C6}
\definecolor{myyellow}{HTML}{FFFADF}
\definecolor{myred}{HTML}{FF0000}

\def\ie{\emph{i.e.}}
\def\eg{\emph{e.g.}}
\def\etal{{\em et al.}}
 
\usepackage{rotating}

\usepackage{xcolor}
\definecolor{citegreen}{RGB}{34,139,34}

\definecolor{iccvblue}{rgb}{0.21,0.49,0.74}
\usepackage[pagebackref,breaklinks,colorlinks,allcolors=iccvblue]{hyperref}

\def\paperID{***} % *** Enter the Paper ID here
\def\confName{ICCV}
\def\confYear{2025}

\title{Medical World Model:\\ Generative Simulation of Tumor Evolution for Treatment Planning}

\author{
    Yijun Yang$^{1,}$\footnotemark[1], 
    Zhao-Yang Wang$^{2}$,
    Qiuping Liu$^{3}$,  
    Shuwen Sun$^{3}$,
    Kang Wang$^{4}$,
    Rama Chellappa$^{2}$,\\
    Zongwei Zhou$^{2}$,
    Alan Yuille$^{2}$,
    Lei Zhu$^{1,5,}$\footnotemark[2],
    Yu-Dong Zhang$^{3}$,
    Jieneng Chen$^{2,}$\footnotemark[2]
    \\
    $^1$The Hong Kong University of Science and Technology (Guangzhou)~~~
    $^2$Johns Hopkins University~~~\\
    $^3$The First Affiliated Hospital of Nanjing Medical University\\
    $^4$University of California, San Francisco
    $^5$The Hong Kong University of Science and Technology\\
{\tt\small Project page: \url{https://yijun-yang.github.io/MeWM}}
}

\begin{document}
\maketitle

\renewcommand{\thefootnote}
{\fnsymbol{footnote}} \footnotetext[1]{Work done while visiting at JHU.}

{\fnsymbol{footnote}} \footnotetext[2]{Corresponding authors.}

\doparttoc % Tell to minitoc to generate a toc for the parts
\faketableofcontents
%%%%%%%%% ABSTRACT 
\begin{abstract}
Providing effective treatment and making informed clinical decisions are essential goals of modern medicine and clinical care.
We are interested in simulating disease dynamics for clinical decision-making, leveraging recent advances in large generative models.
To this end, we introduce the Medical World Model (MeWM), the first world model in medicine that visually predicts future disease states based on clinical decisions. 
MeWM comprises (i) vision-language models to serve as policy models, and (ii) tumor generative models as dynamics models. The policy model generates action plans, such as clinical treatments, while the dynamics model simulates tumor progression or regression under given treatment conditions.  
Building on this, we propose the inverse dynamics model that applies survival analysis to the simulated post-treatment tumor, enabling the evaluation of treatment efficacy and the selection of the optimal clinical action plan. As a result, the proposed MeWM simulates disease dynamics by synthesizing post-treatment tumors, with state-of-the-art specificity in Turing tests evaluated by radiologists. 
Simultaneously, its inverse dynamics model outperforms medical-specialized GPTs in optimizing individualized treatment protocols across all metrics.
Notably, MeWM improves clinical decision-making for interventional physicians, boosting F1-score in selecting the optimal TACE protocol by 13\%, paving the way for future integration of medical world models as the second readers.
% paving the way for future advancements in clinical practice.

\end{abstract}

%%%%%%%%% BODY TEXT
\section{Introduction}
\label{sec:intro}
\begin{figure}[ht!]
\centering
\includegraphics[width=\linewidth]{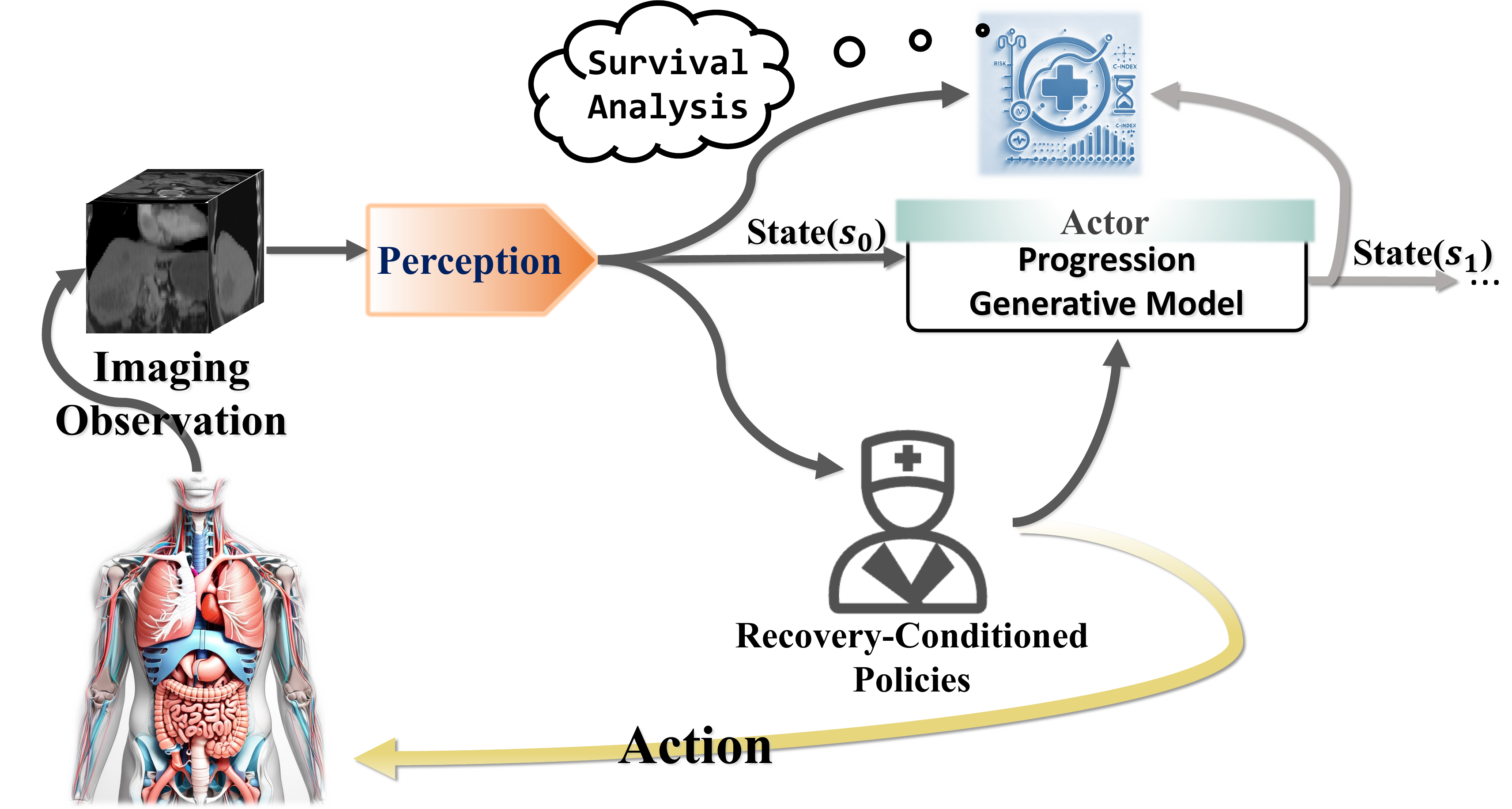}
\vspace{-4mm}
\caption{\textbf{Formulation of Medical World Model.}
It integrates imaging observations with perception modules to form an initial state, which is then processed by a progression generative model to predict future states of disease under different treatment conditions. Recovery-conditioned policies guide treatment decisions, creating a feedback loop for optimizing clinical interventions.
}
\label{fig:banner}
\vspace{-6mm}
\end{figure}

% Qeustion: the relationship of (1) clinical decision making (2) treatment (3) progonosis
% one paragraph: significance of medical treatment

% what's the current gap ()

% how to address the gap?

Clinical decision-making is at the heart of patient care, driving outcomes and shaping the trajectory of healthcare interventions. Physicians constantly weigh the multimodal factors including medical images and patient history to determine the best course of action for each patient. Artificial intelligence (AI) models are increasingly assuming a crucial role in this process by analyzing the complex multimodal data, revealing patterns that may be difficult to detect with human observation alone, and suggesting tailored treatment strategies based on predictive analytics.

Foundation models~\cite{bommasani2021opportunities} such as large language models (LLMs)~\cite{hurst2024gpt, team2023gemini} present a new frontier in medical AI research and development. However, recent studies~\cite{hager2024evaluation} show that LLMs, even those specifically tailored for medicine~\cite{singhal2023large, chen2023meditron, singhal2025toward}, diagnose significantly worse than clinicians and make less informed treatment decisions. This is due to several challenges. 
\textit{\textbf{First}}, the complexity of diseases themselves, such as tumors that evolve under the influence of diverse biological and chemotherapy factors, calls for models that can adapt and account for disease progression. 
\textit{\textbf{Second}}, clinical decision-making necessitates not only accurate predictions but also visually trackable insights that physicians can trust.

Recent breakthroughs in world models (WMs)~\cite{ lecun2022path, bruce2024genie, lu2024generative, wang2024drivedreamer, bar2025navigation} provide a promising avenue for overcoming these obstacles. By generating a predictive distribution of how the world states evolve, WMs mirror the way human planners imagine future scenarios and then make informed decisions via inverse dyanmics~\cite{du2024video, bar2025navigation}. Although they remain largely underexplored in the medical domain, world models hold significant potential for generating clinically realistic images and simulating disease progression, which in turn can facilitate more effective and visually trackable treatment planning.
Figure~\ref{fig:banner} illustrates our formulation of introducing WMs into generalized medical scenarios and how WMs integrate these capabilities to support clinical decision-making.

In this work, we introduce Medical World Model (MeWM) to address these challenges and push the boundaries of AI-driven clinical decision support. 
MeWM comprises three primary components: 
(1) \textbf{a Policy Model} powered by vision-language architectures, which generates the potential action combos from a patient’s current state and specific clinical scenario; 
(2) \textbf{a Dynamics Model} that forwards and simulates tumor dynamics, predicting how tumors could progress or regress under different treatment conditions by generative modeling; 
(3) \textbf{an Inverse Dynamics Model} that performs survival risk analysis on the simulated post-treatment tumor, and quantitatively evaluates treatment efficacy. Beyond forward simulation, this system heuristically explores the optimal plan with the assistance of a segmentation model.
% (4) \textbf{an assistant model}, which achieves the automatic segmentation of the target tumor for better optimization.
% Thus, beyond forward simulation, this system formulates an inverse dynamics model that performs survival analysis on the simulated post-treatment tumor, effectively quantifying treatment efficacy and guiding the selection of the most promising intervention. 
By uniting these elements, MeWM delivers a holistic framework for decision-making: it can synthesize realistic post-treatment tumors that pass Turing tests against radiologists, and it outperforms specialized GPT-like models on Transarterial Chemoembolization (TACE) Protocol Exploration (over 10\%$\uparrow$ in F1-score).

Overall, our contributions are threefold. 
% \jc{TODO: revise the wording}
\begin{itemize}
    \item We propose the medical world models, where we develop a multimodal policy model that leverages vision-language capabilities to propose a tailored set of treatment action combos, and we design a generative dynamics model that accurately captures potential evolution of tumors, enabling forward-looking simulations for different interventions.
    \item We integrate an inverse dynamics model that translates these action-conditioned simulation into survival analysis metrics, thereby offering a transparent and evidence-based tool for choosing the optimal treatment protocol.
    \item We demonstrate a substantial leap in AI-driven decision support for interventional medicine, improving the F1-score in selecting the optimal treatment protocol by 13\% and offering a compelling glimpse into the future of precision healthcare.
\end{itemize}

\section{Related Work}
\label{sec:related}
\subsection{Generative World Modeling}
World models~\cite{ha2018world,lecun2022path} aim to simulate dynamic environments by predicting future states and rewards based on current observations and actions. 
Originally developed for constrained settings like Atari games~\cite{hafner2023mastering}, their ability to model state transitions has been extended to real-world scenarios through joint learning of policies and world models, improving sample efficiency in simulated robotics~\cite{seo2023masked}, real-world robots~\cite{wu2023daydreamer} and autonomous driving~\cite{hu2023gaia,wang2023towards}. 
While early world models focused on simple state transitions, modern approaches integrate structured action-object relationships~\cite{tulyakov2018mocogan} and multi-modal conditioning~\cite{blattmann2023stable,girdhar2023emu}. 
For instance, Du~\etal~\cite{du2023video} present long-horizon video plans by synergising vision-language models and text-to-video models.
Luo~\etal~\cite{luo2024grounding} propose to ground video models to continuous action by leveraging video-guided goal-conditioned exploration to learn a goal-conditioned policy. 
In embodied decision-making, Lu~\etal~\cite{lu2024generative} enables agents to imaginatively explore the world with high generation quality and exploration consistency using video generative models. 
However, there is still no work investigating the applicability of world modeling in medical image analysis and clinical decision-making.

\subsection{Tumor Synthesis}
Tumor synthesis has emerged as an attractive research topic across various medical imaging modalities, such as CT~\cite{yao2021label,lyu2022pseudo,chen2024towards}, MRI~\cite{huang2022common,billot2023synthseg,xing2024cross}, and endoscopic videos~\cite{li2024endora,chen2024surgsora}.
There are also many works on synthesizing non-cancerous lesions including chest CT synthesis~\cite{lyu2022pseudo,yao2021label,bluethgen2024vision,yao2025addressing}, and diabetic lesion synthesis in retinal images~\cite{costa2017end,zhou2019high}. 
Recent studies focus on improving the realism of synthetic tumors in the liver, kidney and pancreas~\cite{hu2023label,lai2024pixel,chen2024towards} by leveraging the large generative models like diffusion models~\cite{ho2020denoising,sohl2015deep, rombach2022high}. 
% AI trained on these synthetic tumors perform similarly well as those trained with real tumors. 
While these methods are conditioned only on shape masks, Li~\etal~\cite{li2024text} propose text-driven tumor synthesis by descriptive reports and conditional diffusion models.
However, most of these works implemented tumor synthesis as a data augmentation to improve tumor  detection tasks. They overlook its potential to empower clinical decision-making in treatment planning.
In this work, we delve into the relatively unexplored field of tumor dynamics simulation by generating post-treatment tumors using pre-treatment scans and treatment actions.

\begin{figure*}[t]
\centering
\includegraphics[width=0.9\textwidth]{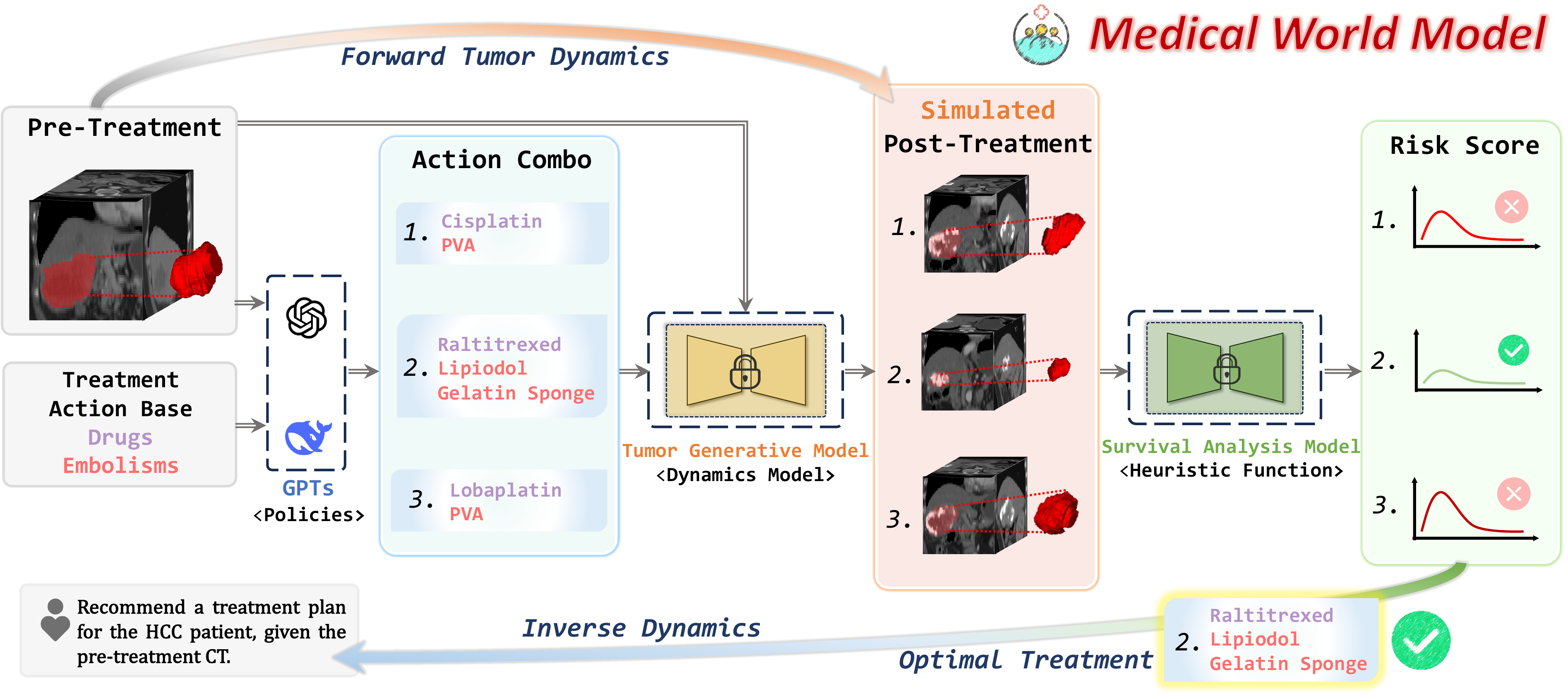}
\vspace{-2mm}
\caption{\textbf{Overview of TACE Protocol Exploration by Medical World Model.}
(1) GPTs (Policy Model): construct the TACE action combos by the observation of pre-treatment CT, integrating clinical guidelines and policies.
(2) Tumor Generative Model (Dynamics Model): simulates post-treatment tumor based on different TACE intervention protocols, predicting treatment outcomes.
(3) Survival Analysis Model (Heuristic Function): assesses risk scores from both simulated post-treatment CT and pre-treatment CT to determine the most effective TACE protocol.
Note that the 3D tumor masks (colored in red) can be extracted using a well-trained segmentation network (as Assistant Model).
The framework enables visually trackable protocol optimization by iterating between clinical policy guidance, generative modeling, and survival analysis.
}
\label{fig:framework}
\vspace{-4mm}
\end{figure*}

\subsection{Prognosis and Clinical Decision-making}
Post-treatment prognosis in medical imaging is essential for evaluating therapy effectiveness, predicting disease recurrence, and guiding further clinical decisions. CT is widely used to assess structural and functional changes in tumors following interventions such as surgery, chemotherapy, radiation therapy, transarterial chemoembolization (TACE), and immunotherapy~\cite{lee2024deep,huang2025ct,hagag2024deep,liang2023deep,yao2020deepprognosis,yao2021deepprognosis}. 
Lee~\etal~\cite{lee2024deep} employ a CNN-based model to predict the post-treatment survival of patients with hepatocellular carcinoma (HCC) using CT images and clinical information. 
In addition, LLMs
% , such as GPT-based models, 
are increasingly being explored to assist in clinical decision-making \cite{hager2024evaluation,li2025large,busch2025multilingual,kim2024mdagents}. However, little attention has been given to applying LLMs for post-treatment prognosis.
They did not leverage the feedback from survival analysis to achieve prompt intervention as well.

\section{Medical World Models}
\label{sec:method}

\noindent\textbf{Overall Framework}.
As shown in Fig.~\ref{fig:framework}, our MeWM takes a visual observation of pre-treatment CT $x_0$, a language treatment goal $g$ to simulate the future state and explore the best treatment protocol.
\textbf{Policy model} (\S~\ref{sec:method:policy}) acquires the descriptive observation based on the visual state, and constructs a set of treatment protocols by the language goal and clinical guidelines.
To perform the exploration, given the pre-treatment CT and an action combo, the \textbf{dynamics model} (\S~\ref{sec:method:dynamics}) predicts the concrete resulting state, \ie, generating post-treatment CT.
Finally, \textbf{inverse dynamics model} (\S~\ref{sec:method:inverse}) driven by Heuristic Function predicts the risk score from pre-treatment CT and simulated post-treatment CT with tumor masks from Assistant Model, to effectively prune branches in search and heuristically determine the optimal solution.

\subsection{Policy Model}
\label{sec:method:policy}
Vision-language models~\cite{chen2024huatuogpt,zheng2024large} have emerged as a powerful source of prior knowledge about the clinical world, providing rich information about how to complete promising treatment from large-scale internet data and clinical guidelines.
Based on TACE clinical guidelines, we set up the exploration configurations, including all potential chemotherapy drugs (\eg, Raltitrexed, Cisplatin) and embolism materials (\eg, Lipiodol, Gelatin Sponge).
The two parts constitute the action base, which provides possible TACE protocols for Generative Dynamics Models as conditions.
Then, we adopt a pre-trained large multimodal model (LMM), \eg, GPT-4o, to serve as policies.
Given a high-level goal $g$ (\eg, ``What TACE treatment protocols are recommended for a patient diagnosed with hepatocellular carcinoma (HCC) given the pre-treatment CT?''), the policy model $\pi_\text{VLM}(x_0,g)$ extracts the visual observation and tumor context from the given pre-treatment CT $x_0$ to prompt the proper Transarterial Chemoembolization (TACE) actions.
To constrain the excessively large tree search in the action base, we further prompt the Large Language Reasoning Model, \ie, Deepseek-R1~\cite{guo2025deepseek}, to refine the drug set and embolism set by the clinical policies, whose final cardinalities are $D$ and $E$, respectively.

\begin{figure*}[t]
\centering
\includegraphics[width=0.75\textwidth]{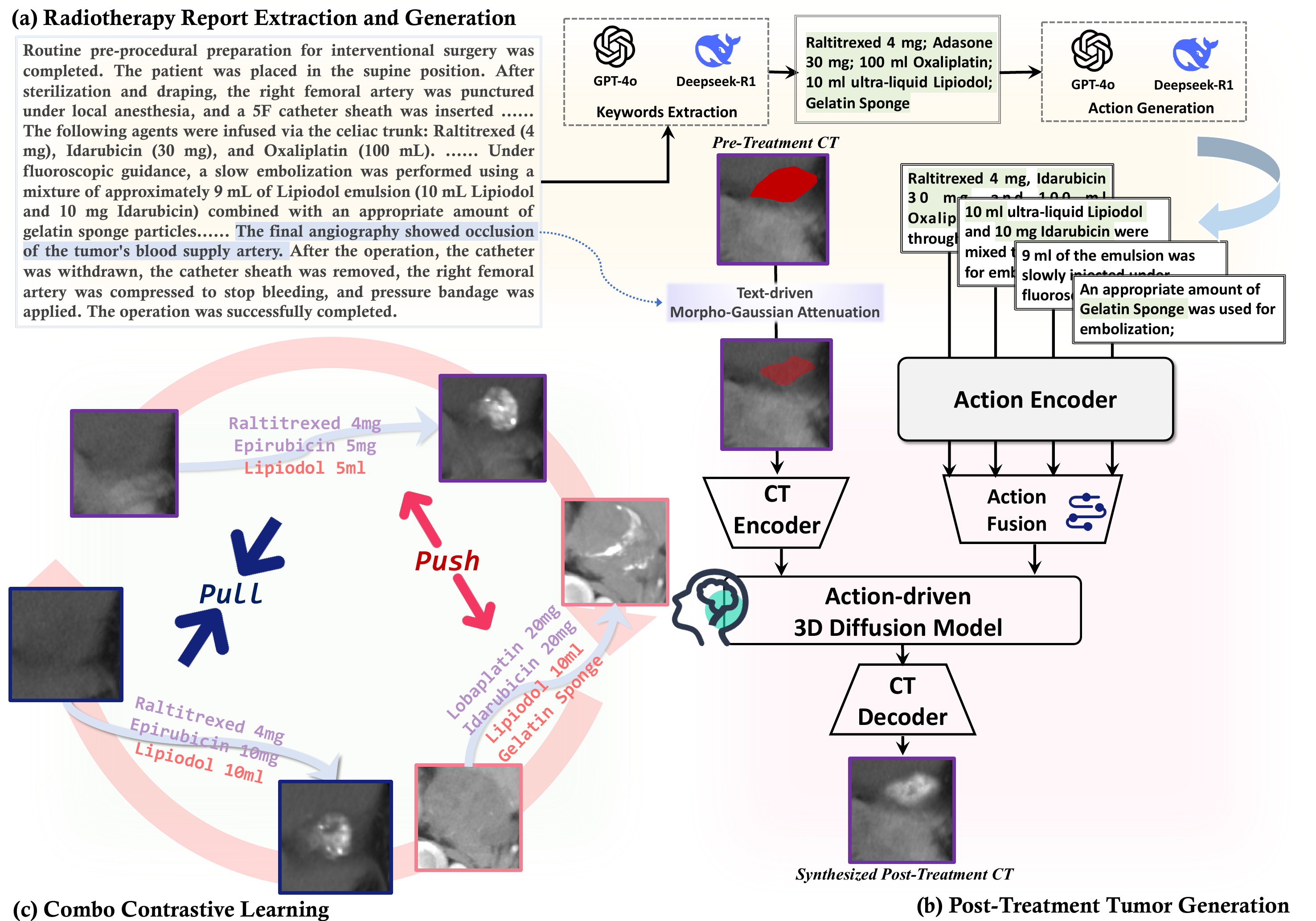}
\vspace{-2mm}
\caption{\textbf{Dynamics Model based on Tumor Generative Model.} The training framework consists of three parts:
\textit{(a) Radiotherapy Report Extraction and Generation}: GPT-4o and Deepseek-R1 extract key treatment details from radiotherapy reports and generate corresponding TACE surgical actions.
\textit{(b) Post-Treatment Tumor Generation}: An Action-driven 3D Diffusion Model is conditioned by fused action embeddings and attenuated CT features to generate post-treatment tumors that simulate treatment outcomes.
\textit{(c) Combo Contrastive Learning (CCL)}: The model learns from treatment variations by pushing apart dissimilar combos and pulling together similar ones, improving its ability to generate realistic and action-aware post-treatment tumor appearances.
}
\label{fig:generative}
\vspace{-4mm}
\end{figure*}

\subsection{Dynamics Model}
\label{sec:method:dynamics}
\noindent\textbf{Radiotherapy Report Extraction and Generation.}
While most existing studies focus on human-authored radiology reports, we aim to address radiotherapy reports to extract more comprehensive information on treatment protocols. However, raw radiotherapy reports pose significant challenges due to noise and fragmented information, which hinder controlled tumor synthesis.
To mitigate these issues, we propose a two-stage text preprocessing framework consisting of data cleaning and augmentation. In the first stage, we perform keyword extraction by aggregating the outputs of both GPT-4o and DeepSeek-R1, focusing on key entities such as drugs, embolic agents, and their corresponding dosages.
In the second stage, we leverage the same tools for text generation, constructing a structured core action description based on the extracted keywords. 
This approach enhances the consistency and informativeness of processed reports, facilitating downstream tasks in tumor synthesis and treatment analysis.

\noindent\textbf{Post-Treatment Tumor Generation. }
We adopt Latent Diffusion Models (LDMs)~\cite{rombach2022high} for latent feature extraction from 3D Pre-treatment CT volumes and integrate textual action embedding for controlled tumor synthesis.
Each 3D Post-treatment CT volume $x_1\in\mathbb{R}^{H\times W\times D}$ is encoded into a lower dimensional latent representation $z_1 = \mathcal{E}(x_1)$ using a 3D VQGAN autoencoder~\cite{esser2021taming}. 
In the latent space, following the spirit of DiffTumor~\cite{chen2024towards}, we define a diffusion process that progressively adds noise to the latent representation $z_1$ over discrete time steps $t=1,...,T$.
Given a pair of tumor-present pre-treatment CT volume $x_0$ and the mask of its tumor region $m_0$, we condition the denoising model on the masked pre-treatment latent representation $z_0^{'} = \mathcal{E}(m_0^{'} \odot x_0)$ where $m_0^{'}$ is the attenuated mask from $m_0$ by our proposed \textit{Text-driven Morpho-Gaussian Attenuation}.
Specifically, to mimic the effects of TACE treatment, the process begins with occlusion assessment on radiotherapy reports. 
The textual descriptions (\eg, \textit{occluded}, \textit{reduced}, \textit{disappear}) are extracted and analyzed to determine the attenuation level $l \in \{1,2,3,4\}$, where a higher value corresponds to better tumor curative effects.
Then, morphological erosion and dilation with the adaptive kernel by $l$ are applied to $m_0$, simulating occlusion-induced tumor structural dynamics.
Simultaneously, adaptive Gaussian blurring is employed to exhibit the characteristics of heterogeneous intensity changes due to lipiodol deposition, necrotic transformation, and reduced perfusion.
The final attenuated mask $m_0^{'}$ is computed by the three steps, ensuring a smooth transition between tumor and organ tissues. Note that this attenuation is only used during training.

Also, we condition the denoising model on the generated textual action.
Given the action combo $a = \{a_1,...,a_H\}$, each sub-action, respectively, undergoes encoding through a CLIP~\cite{radford2021learning} text encoder $\phi(\cdot)$ followed by linear projection $\sigma_1(\cdot)$, enabling dimension reduction to a latent clinical concept space.
To enhance the semantics of action conditions in $D$ different drug and $E$ embolism keywords, we introduce learnable concept embeddings $c$, which extract keyword representations from the given action combo.
This explicit pharmaceutical grounding enables precise modeling of therapeutic components while maintaining robustness to context variations.
The final action condition $\tau(a)$ is the fusion of holistic text embeddings and concept embeddings by fully connected layers $\sigma_2$, :
\begin{equation}
    \tau(a) = \sigma_2([[\sigma_1(\phi(a_1)),...,\sigma_1(\phi(a_H))],c]),
\end{equation}
where $[\cdot]$ denote concatenation operation.

The training objective of diffusion model is as follows:
\begin{equation}
    \mathbb{E}_{z_1, \boldsymbol{\epsilon} \sim \mathcal{N}(0,1), t} 
\left[ \left\| \boldsymbol{\epsilon} - \epsilon_{\theta} 
\left( z_t, z_0^{'}, m_0, \tau(a), t \right) 
\right\|_2^2 \right],
\end{equation}
where $\epsilon_\theta(\cdot,t)$ is a 3D U-Net with interleaved self-attention layers and convolutional layers~\cite{ho2020denoising,nichol2021improved,chen2024towards} that predict the noise given the input variable and conditions.

\noindent\textbf{Combo Contrastive Learning. }
We adopt a contrastive learning strategy that aligns action combos with tumor evolution to enhance the realism and discrimination of post-treatment tumor synthesis. 
Given a pre-treatment anchor pair $(x_0,m_0)$, along with an action combo $a$, the goal is to generate a post-treatment CT $\hat{x}$ by generative model $f_\text{DM}(\cdot)$.

For each anchor, positive samples, $\hat{x}^+ = f_\text{DM}(x_0^+,m_0^+,a^+)$, are defined as synthetic post-treatment CT from another pair but its action combo contains the same drug/embolism keywords. 
Negative samples, $\hat{x}^- = f_\text{DM}(x_0,m_0,a^-)$, in contrast, are generated using the same pair but action combos with diverse keywords, leading to distinct tumor evolution patterns. 
This contrastive loss is incorporated into the tumor generative model: 
\begin{equation}
    \mathbb{E} \left[ -\log \frac{\exp \left( \text{sim}(\hat{x}, \hat{x}^+) / \delta \right)}
    {\sum \exp \left( \text{sim}(\hat{x}, \hat{x}^-) / \delta \right) - \exp \left( \text{sim}(\hat{x}, \hat{x}^+) / \delta \right)} \right],
\end{equation}
where $\text{sim}(\cdot)$ denotes cosine similarity, $\delta$ is the temperature scaling factor.
This contrastive learning strategy ensures that tumors simulated from similar treatment protocols exhibit consistent attenuation effects, while those derived from distinct protocols remain differentiable.
%

% \vspace{-2mm}
\subsection{Inverse Dynamics Model}
\label{sec:method:inverse}
Inverse Dynamics Model, which empowers our full framework, aims to infer the most effective treatment strategy by analyzing relationships between pre-treatment conditions, intervention actions, and expected post-treatment outcomes.
We unfold its essence in three aspects: (1) Assitant Model; (2) Heuristic Function and (3) TACE Protocol Exploration.

\noindent\textbf{Assistant Model. }
To better discriminate the tumor in synthesized post-treatment CT $\hat{x}$, 
we introduce a tumor segmentation model as Assistant Model $H_\text{seg}(\cdot)$.
Post-treatment tumors are characterized by heterogeneous high-intensity regions due to \textit{calcification/lipiodol} deposition, irregular shapes reflecting \textit{necrotic tissue} changes, and reduced or absent contrast enhancement in viable tumor areas, in contrast to traditional pre-treatment CT tumors.
Thus, we adapt a pre-trained nnUNet-based~\cite{isensee2021nnu} model to this post-treatment context by finetuning it on our ground truth pairs of post-treatment CT and mask.
Given the well-trained Assistant Model, the simulated post-treatment CT $\hat{x}$ from Tumor Generative Model is processed for the segmentation of liver and tumor.
The post-treatment CT with the predicted mask $\hat{m}$ is subsequently utilized for survival analysis.

\noindent\textbf{Heuristic Function. }
We use survival analysis model to implement a heuristic function $H_\text{surv}(x_0, m_0, \hat{x}, m_0, g)$, which quantifies the efficacy under the specific TACE action combo by the output risk score.
Inspired by DeepSurv~\cite{katzman2018deepsurv}, we utilize a 3D convolution-based model structure, the 3D ResNet (MC3)~\cite{tran2018closer}, as the feature extractor of survival analysis model.
Given the pre-treatment pair $(x_0,m_0)$ and simulated post-treatment pair $(\hat{x}, \hat{m})$, we extract their concatenated CT and mask, respectively, and bidirectionally align the semantics of pre- and post-treatment by Cross-Attention Transformer~\cite{kirillov2023segment}.
After that, we adopt an attention-based aggregator to fuse pre- and post-features, followed by fc layers to determine the risk score.
The action combo with a lower risk score should bring greater efficacy for the patient.
Note that, for training, we leverage multi-task learning strategy, \ie, CoxPH~\cite{katzman2018deepsurv} and OS regression, to improve the generalization of survival analysis.

\begin{figure*}[t]
\centering
\begin{minipage}{0.95\textwidth}
\begin{algorithm}[H]
    \begin{algorithmic}[1]
    \STATE \small{\textbf{Input:} Pre-treatment CT $x_0$, Pre-treatment tumor mask $m_0$, Language treatment goal $g$}
    \STATE \small{\textbf{Functions:} VLM Policy Model $\pi_{\text{VLM}}$, Dynamics Model $f_{\text{DM}}$, Heuristic Function $H_{\text{surv}}$, Assistant Model $H_{\text{seg}}$} 
    \STATE \small{\textbf{Hyperparameters:} Drug Actions factor $D$, Embolism Actions factor $E$, Tumor Generative factor $T$, Protocol Beams $B$, Drug horizon $H_d$, Embolism horizon $H_e$} \\
    \STATE plans $\leftarrow [ \hspace{0.1em} [x_0] \hspace{0.5em} \forall \hspace{0.5em} i \in \{1 \ldots B\}]$ \hspace{4.50cm} \small{\color{gray}{\# Initialize B Different TACE Protocol Beams}}
    \STATE $drug_{1:D}, embo_{1:E}, rule \gets \pi_\text{VLM}(x_0, g)$ \hspace{2cm} \small{\color{gray}{\# Generate $D$ Different Drug, $E$ Embolism Actions, Clinical Rules}}
    \FOR{$h = 1 \ldots H_d$}
        \FOR{$b = 1 \ldots B$}
            \STATE tumors $\gets [f_{\text{DM}}(x, drug_i)$ for j in $(1 \ldots T)$ for i in $(1 \ldots D)$ if $rule]$\hspace{0.1cm} \small{\color{gray}{\# Generate tumors from $x$ and plans[b] under rule}}
            \STATE plans[b].append($argmin$(tumors, $H_{\text{surv}}, H_{\text{seg}}$)) \hspace{1cm} \small{\color{gray}{\# Add  Tumor with Lowest Risk to Plan}}
        \ENDFOR
        \STATE max\_idx, min\_idx $\gets$ $argmax$(plans, $H_{\text{surv}}, H_{\text{seg}}$),  $argmin$(plans, $H_{\text{surv}}, H_{\text{seg}}$)
        \STATE plans[max\_idx] $\gets$ plans[min\_idx] \hspace{2.55cm} \small{\color{gray}{\# Periodically Replace the Plan with High Risk}}
    \ENDFOR
    \FOR{$h = 1 \ldots H_e$}
        \FOR{$b = 1 \ldots B$}
            \STATE tumors $\gets [f_{\text{DM}}(x, embo_i)$ for j in $(1 \ldots T)$ for i in $(1 \ldots E)$ if $rule]$\hspace{0.1cm} \small{\color{gray}{\# Generate tumors from $x$ and plans[b] under rule}}
            \STATE plans[b].append($argmin$(tumors, $H_{\text{surv}}, H_{\text{seg}}$)) \hspace{1cm} \small{\color{gray}{\# Add Tumor with Lowest Risk to Plan}}
        \ENDFOR
        \STATE max\_idx, min\_idx $\gets$ $argmax$(plans, $H_{\text{surv}}, H_{\text{seg}}$),  $argmin$(plans, $H_{\text{surv}}, H_{\text{seg}}$)
        \STATE plans[max\_idx] $\gets$ plans[min\_idx] \hspace{2.55cm} \small{\color{gray}{\# Periodically Replace the Plan with High Risk}}
    \ENDFOR
    \STATE plan $\gets$ $argmin$(plans, $H_{\text{surv}}, H_{\text{seg}}$) \hspace{5.6cm} \small{\color{gray}{\# Return Plan with Lowest Risk}}
    \end{algorithmic}
    \caption{\small TACE Protocol Exploration with MeWM}
    \label{alg:cond_gen}
\end{algorithm}
\end{minipage}
% \vspace{-4mm}
\end{figure*}

\noindent\textbf{TACE Protocol Exploration. }
Given a combination of the proposed models above, we are able to predict TACE protocol from any Pre-treatment CT $x$ by a language treatment goal $g$.
To reason the optimal action combo, we propose to search for a list of actions to reach $g$, corresponding to finding a treatment plan consisting of both drug and embolism components, which optimizes:
\begin{equation}
    \hat{x}^{*}_{1:H} = \mathop{\arg\min}\limits_{\hat{x}_{1:H}\sim\pi_\text{VLM},f_\text{DM}} H_{\text{surv}}(x_0, m_0, \hat{x}, 
    H_\text{seg}(\hat{x}), g), 
\end{equation}
where $H = H_d+H_e$.

With this objective in mind, we exhibit a tree-search exploration procedure.
Our exploration algorithm initializes a set of $B$ parallel protocol beams.
We sample the potential action space composed of $D$ drugs and $E$ embolisms and clinical rules using $\pi_\text{VLM}$.
The clinical rules are introduced to prune unreasonable branches, \eg, concomitant use of multiple platinum-based agents is contraindicated due to the risk of cumulative toxicity and myelosuppression.
We sequentially explore the two parts to ensure that TACE protocol contains both drugs and embolism.
For each current action combo, we synthesize $T$ post-treatment tumors from $f_\text{DM}(x_0,m_0,a)$ to obtain a more reliable simulation.
We then use our heuristic function $H_\text{surv}(x_0, m_0, \hat{x},\hat{m},g)$ with the assistance of $H_\text{seg}(\hat{x})$ to select the generated tumor with the best average survival score of $T$ replicas among $D$ or $E$ actions.
After every step of extending all beams, we discard the beam with the worst survival score and replace its action combo with the best beam.
To prevent cumulative toxicity and organ dysfunction, we prohibit over-exploration by drug horizon $H_d$ and embolism horizon $H_e$.
Our final action combo is taken from the beam with the best survival score and adopted as TACE protocol for the patient.
The pseudocode of our method is also provided in Algorithm~\ref{alg:cond_gen}.

\vspace{-2mm}
\section{Experiments}
\label{sec:exp}

\begin{table*}[t]
\renewcommand{\arraystretch}{1.15}
  \centering
  \caption{\textbf{Action-driven Visual Turing Test} involves three radiologists (R1-R3) each evaluating five groups of 48 CT scans each, with 24 real post-treatment tumors and 24 synthetic post-treatment tumors from a tumor generative model, respectively. They were tasked with categorizing each CT scan as either real or synthetic. A higher sensitivity score indicates better discriminative ability of radiologists, while a lower specificity score indicates a higher number of synthetic tumors being identified as real. We also provide perceptual evaluation using FID and LPIPS compared to corresponding real post-treatment scans. Lower FID and LPIPS indicate better simulation results.}
  \vspace{-3mm}
  \resizebox{0.9\linewidth}{!}{%
\begin{tabular}{c|ccc|ccc|ccc|cc} 
\toprule
\multirow{2}{*}{Methods} & \multicolumn{3}{c|}{\textbf{R1}}              & \multicolumn{3}{c|}{\textbf{R2}}              & \multicolumn{3}{c|}{\textbf{R3}}              & \multicolumn{2}{c}{\textbf{Perceptual metrics}}  \\ 
\cline{2-12}
& \textcolor{gray}{sensitivity} & specificity $\downarrow$ & \textcolor{gray}{accuracy} & \textcolor{gray}{sensitivity} & specificity $\downarrow$ & \textcolor{gray}{accuracy} & \textcolor{gray}{sensitivity} & specificity $\downarrow$ & \textcolor{gray}{accuracy} & FID$\downarrow$  & LPIPS$\downarrow$       \\ 
\hline
SynTumor~\cite{hu2023label}                            & \textcolor{gray}{100.0}         & 95.83       & \textcolor{gray}{97.92}    & \textcolor{gray}{87.50}        & 95.83       & \textcolor{gray}{91.67}    & \textcolor{gray}{100.0}         & 95.83       & \textcolor{gray}{97.92}    & 3.33 & 0.6832                           \\
Pixel2Cancer~\cite{lai2024pixel}                        & \textcolor{gray}{95.83}       & 100.0         & \textcolor{gray}{97.92}    & \textcolor{gray}{91.67}       & 95.83       & \textcolor{gray}{93.75}    & \textcolor{gray}{100.0}         & 100.0         & \textcolor{gray}{100.0}      & 3.34 & 0.6831                           \\
DiffTumor~\cite{chen2024towards}                           & \textcolor{gray}{100.0}         & 91.67       & \textcolor{gray}{95.83}    & \textcolor{gray}{95.83}       & 87.50        & \textcolor{gray}{91.67}    & \textcolor{gray}{100.0}         & 87.50        & \textcolor{gray}{93.75}    & 1.40  & 0.7660                            \\
TextoMorph~\cite{li2024text}                          & \textcolor{gray}{100.0}         & 91.67       & \textcolor{gray}{95.83}    & \textcolor{gray}{91.67}       & 83.33       & \textcolor{gray}{87.50}     & \textcolor{gray}{95.83}       & 87.50        & \textcolor{gray}{91.67}    & 1.03 & 0.9111                           \\ 
\hline
\rowcolor{blue!7!white} MeWM (Ours)                          & \textcolor{gray}{100.0}         & \textbf{79.17}       & \textcolor{gray}{89.58}    & \textcolor{gray}{91.67}       & \textbf{70.83}       & \textcolor{gray}{81.25}    & \textcolor{gray}{91.67}       & \textbf{75.00}          & \textcolor{gray}{83.33}    & \textbf{0.71} & \textbf{0.6120 }                           \\
\bottomrule
\end{tabular}
  }
\vspace{-2mm}
  \label{tab:turing}%
\end{table*}

\noindent\textbf{HCC-TACE In-house Dataset. }
We collect a large repository of 338 longitudinal paired pre- and post-treatment CT scans with well-annotated liver/tumor masks and clinical records, such as TACE radiotherapy reports as gold action and Overall Survival (OS) time.
We split the training set (validation set included) and testing sets in the 9:1 ratio.

\noindent\textbf{HCC-TACE-Seg Public Dataset~\cite{morshid2019machine}. }
For external validation, we use patients from HCC-TACE-Seg public dataset referring to a single-institution collection with confirmed HCC treated at The University of Texas MD Anderson Cancer Center.
We conduct data curation and preprocessing to collect 78 cases containing pre-treatment CT, post-treatment CT, TACE Gold Action, and OS time.
We use 80\% cases to fine-tune and validate MeWM and leave 20\% cases for the exploration evaluation.

\subsection{Evaluation on Generation Quality}
\begin{figure}[t]
\centering
\includegraphics[width=0.85\linewidth]{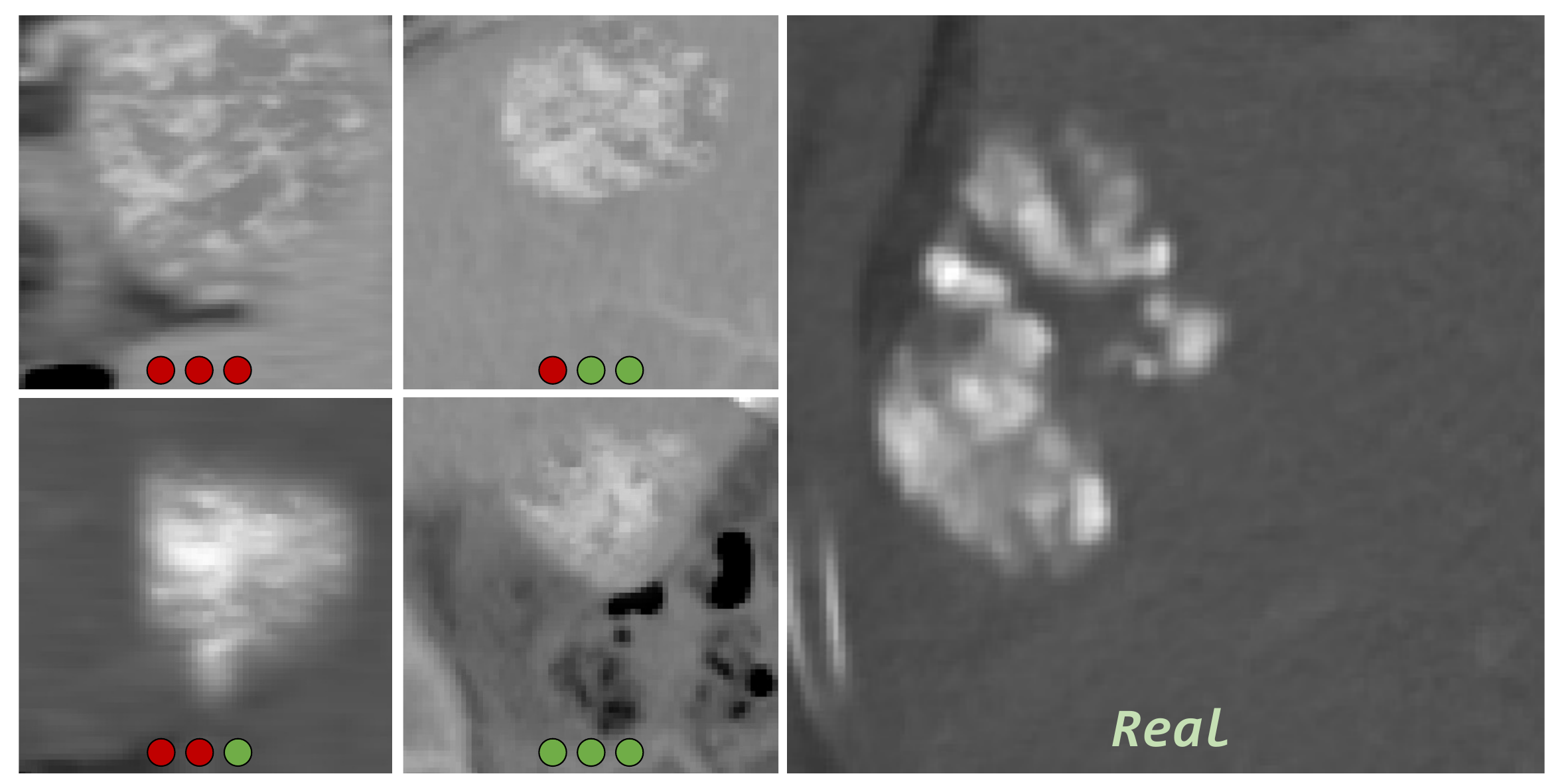}
\vspace{-2mm}
\caption{\textbf{Examples of Visual Turing Test}. We present one real tumor alongside examples of synthetic tumors that were correctly and incorrectly identified. A \textcolor{Mahogany}{red dot} indicates the radiologist classified the post-treatment tumor as synthetic, while a \textcolor{OliveGreen}{green dot} signifies it was identified as real.
}
\label{fig:turing}
\vspace{-4mm}
\end{figure}

\noindent\textbf{Visual Turing Test (Human Evaluation). }
We conduct an action-driven Visual Turing Test on 240 CT scans of post-treatment tumors, where 120 scans contain real post-treatment tumors, and 120 scans contain synthetic post-treatment tumors generated by different tumor synthesis models. Three radiologists (R1-R3) participated in this study, independently evaluating five groups of 48 CT scans each and classifying them as either real or synthetic.
It is important to note that the radiologists' evaluations are based on whether the synthetic tumor closely resembles a post-treatment tumor, which typically contains a mixture of lipiodol deposition, necrotic, and viable tumor regions, distinguishing it from ordinary pre-treatment tumors.
The test results are summarized in Table~\ref{tab:turing}. The sensitivity scores of all radiologists remain high (above 91\%), demonstrating their ability to correctly identify real post-treatment tumors. However, specificity scores vary among the methods, indicating different levels of realism in the synthetic tumors. Notably, our method MeWM achieves the lowest specificity scores (79.17\% for R1, 70.83\% for R2, and 75.00\% for R3), suggesting that a large proportion of synthetic tumors generated by our approach are mistaken as real. This indicates superior realism compared to other methods such as SynTumor~\cite{hu2023label}, Pixel2Cancer~\cite{lai2024pixel}, DiffTumor~\cite{chen2024towards}, and TextoMorph~\cite{li2024text}.
Figure~\ref{fig:turing} illustrates examples from the test, where a real tumor is compared with synthetic tumors that radiologists correctly or incorrectly classified.
This highlights synthetic tumors closely resemble real post-treatment tumors.

\noindent\textbf{Perceptual Evaluation. }
We perform perceptual evaluation using FID and LPIPS scores, where lower values indicate better simulation quality. Our method achieves the best FID (0.71) and LPIPS (0.6120), demonstrating the highest fidelity in synthetic tumor generation. These results confirm that MeWM effectively synthesizes realistic post-treatment tumors, making it more challenging for radiologists to distinguish between real and synthetic cases.

\begin{figure}[t]
\centering
\includegraphics[width=0.9\linewidth]{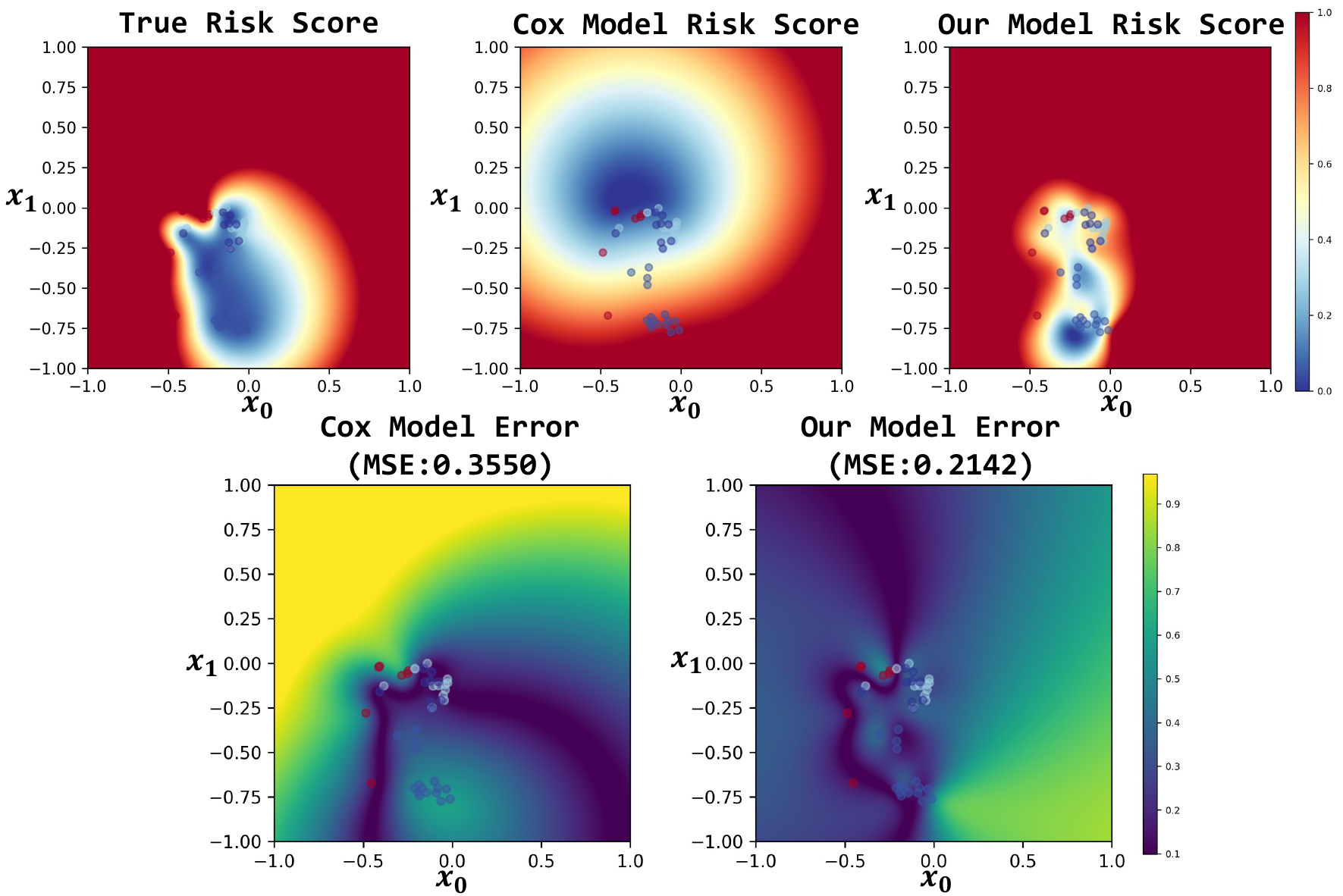}
\vspace{-2mm}
\caption{\textbf{Performance of heuristic function on survival analysis.} The first three heatmaps show the true risk distribution, Cox model predictions, and our heuristic function predictions. The last two depict prediction errors, with lower MSE (0.2142) for our model compared to the Cox model (0.3550), demonstrating improved accuracy in capturing localized risk patterns.
}
\label{fig:surv}
\vspace{-2mm}
\end{figure}

\subsection{Survival Analysis}
\begin{figure}[t]
\centering
\includegraphics[width=\linewidth]{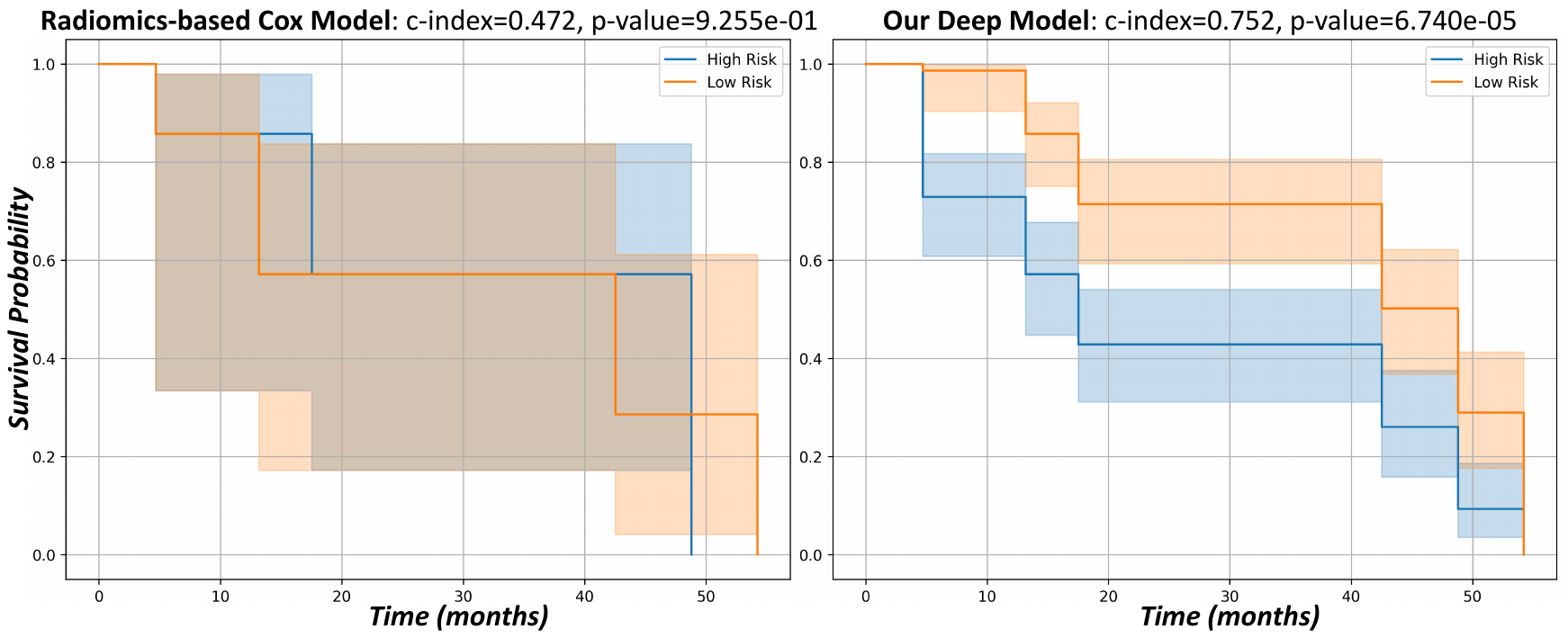}
\vspace{-4mm}
\caption{\textbf{Kaplan-Meier Survival Curves: Radiomics-based Cox Model~\cite{radio_page} vs. our deep model.} 
The left shows the survival curves predicted by the Cox model based on Radiomics features. The right presents the survival curves from our model based on deep features, which achieves a significantly higher c-index of 0.752 and a log-rank p-value of $6.74e-5$, demonstrating a stronger ability to distinguish between high- and low-risk groups. Shaded areas represent confidence intervals.
}
\label{fig:radiomics}
\vspace{-4mm}
\end{figure}

In Figure~\ref{fig:surv}, we evaluate survival risk regression between the popular Cox Proportional Hazards model~\cite{fox2002cox} and our heuristic function model on the HCC-TACE-Seg dataset. 
The true risk distribution (left) is estimated using the Nelson-Aalen estimator~\cite{colosimo2002empirical}. 
The Cox model fails to accurately distinguish between high- and low-risk samples from low-dimensional deep features, resulting in an overly smoothed risk distribution. 
In contrast, our model produces a risk map that better aligns with the true distribution, effectively capturing variations in risk levels.
Error analysis shows higher Mean Square Error (MSE), \ie, 0.3550 for Cox and lower MSE, 0.2142 for our model, indicating superior accuracy. 
Figure~\ref{fig:radiomics} further presents Kaplan-Meier survival curves comparing risk stratification performance between the Radiomics-based Cox model and our deep model.
These results demonstrate that our heuristic function better estimates survival risks, reduces prediction errors, and captures complex patterns beyond the capabilities of the Cox model and radiomics features.

\begin{table*}[t]
\renewcommand{\arraystretch}{1.15}
  \centering
  \caption{\textbf{TACE Protocol Exploration Evaluation on HCC-TACE in-house dataset and public dataset}. F1-score, Jaccard index, Precision, and Recall are computed between the predicted action combo and gold action. MeWM significantly advances multimodal GPTs in exploring optimal individualized treatment protocol across all metrics, even comparable to interventional physicians.
  }
  \vspace{-3mm}
  \resizebox{0.9\linewidth}{!}{%
\begin{tabular}{lcccc|cccc} 
\toprule
   & \multicolumn{4}{c|}{In-house dataset}        & \multicolumn{4}{c}{Public dataset}                 \\ 
 \multirow{-2}{*}{{}\textbf{Methods}}              & F1-score$\uparrow$ & Jaccard$\uparrow$ & Precision$\uparrow$ & Recall$\uparrow$ & F1-score$\uparrow$ & Jaccard$\uparrow$ & Precision$\uparrow$ & Recall$\uparrow$  \\ 
\hline
 Physician w/ Pre-CT                                                 & 48.81     & 38.44  & 46.67    & 54.67   & 71.43     & 63.10  & 66.67    & 78.57     \\
\rowcolor[rgb]{0.898,1,0.918}  Physician w/ MeWM                                                   &    61.51 (+13\%)      &    49.89     &   60.89     &      65.44  &     80.00 (+9\%)      &    73.81      &     76.16       &   85.71      \\ 
\hline
 Qwen2.5-VL~\cite{Qwen2.5-VL}                                              &     37.09    &     24.49    &     34.44      &    41.83     &      47.14    &    34.40      &   53.57        &   42.86      \\
 GPT-4o~\cite{hurst2024gpt}                                                                    & 41.97    & 27.81   & 35.93     & 52.78   & 44.29    & 32.74    & 57.14     & 38.10    \\ 
 Claude-3.7-sonnet~\cite{anthropic2024claude}                                              &     40.93    &     28.55    &     45.83     &    37.78   &      44.76    &    33.81      &   64.29        &   35.71      \\
\hline
 CT2Rep~\cite{hamamci2024ct2rep} &    27.75      &     17.21    &      30.83     & 25.83       &     43.61     &    28.57      &      53.57     &   37.50      
\\
 MedGPT~\cite{moor2023foundation, medgpt_page}                                                              & 37.51    & 25.57   & 32.78     & 45.21  & 47.14    & 40.48    & 50.00        & 45.24    \\
 HuatuoGPT-Vision~\cite{zhang2023huatuogpt}                                                    & 40.13    & 29.08   & 40.11     & 42.28  & 52.62    & 42.26    & 54.76     & 51.19    \\ 
\hline
\rowcolor{blue!7!white} \textbf{MeWM(Ours)}                                                               &     \textbf{52.38}    &    \textbf{38.59}    &     \textbf{63.06}      &             \textbf{46.17}    &    \textbf{64.08}      &    \textbf{48.45}      & \textbf{72.62}   &      \textbf{58.93}       \\

\bottomrule
\end{tabular}
   }
  \label{tab:quantitative_comparison}%
  % \vspace{-4mm}
\end{table*}

\subsection{Results on TACE Protocol Exploration}
\paragraph{Evaluation Strategy. }
For treatment planning evaluation, we utilize four metrics in Table~\ref{tab:quantitative_comparison}:
(1) \textit{F1-score}: harmonizes Precision and Recall, balancing redundancy and omissions;
(2) \textit{Jaccard Index}: measures prediction overlap with gold actions, emphasizing category-level alignment;
(3) \textit{Precision}: reflects recommendation purity, penalizing incorrect or redundant drugs/embolisms;
(4) \textit{Recall}:
captures therapeutic coverage, highlighting critical omissions.

\noindent\textbf{Partial Observation Misleads GPTs. }For Multi-modal Large Language Models (\eg, GPT-4o, MedGPT, HuatuoGPT-Vision), they are prompted with pre-treatment CT slices and allowed to predict the action combo from the given action set.
These inferior results (over -10\% in F1-score) to MeWM demonstrate that it tends to make deficient planning relying solely on vision-language models and their commonsense reasoning. This also validates the necessity of simulation from pre-treatment to post-treatment. 

\begin{figure}[t]
\centering
\includegraphics[width=\linewidth]{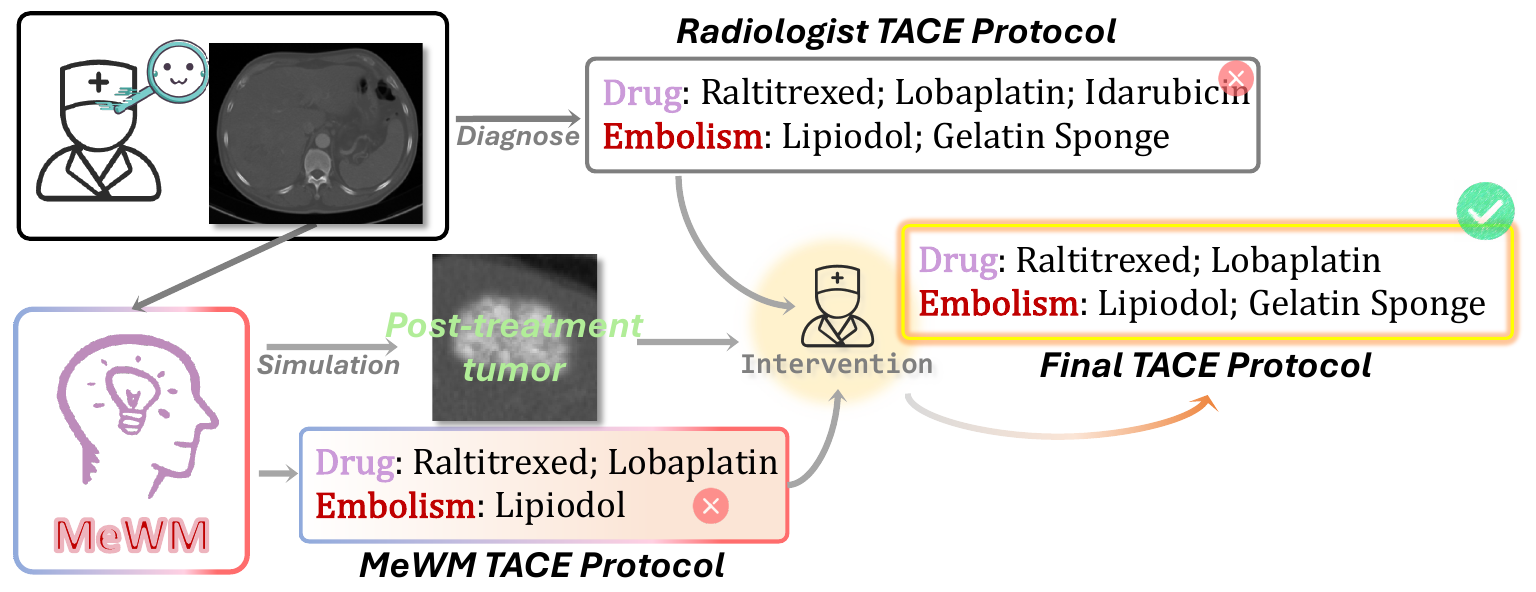}
\vspace{-6mm}
\caption{\textbf{Example of MeWM intervention in clinical applications.}  The radiologist initially proposes a TACE protocol with Raltitrexed, Lobaplatin, Idarubicin, and embolization using Lipiodol and Gelatin Sponge. MeWM simulates a protocol with Raltitrexed, Lobaplatin, and Lipiodol. After intervention, the optimized protocol removes Idarubicin but restores Gelatin Sponge, aligning with gold action.
}
\label{fig:exp_tace}
% \vspace{-4mm}
\end{figure}

\noindent\textbf{MeWM as A Clinical Decision-support Tool. }
MeWM demonstrates significant potential in augmenting the capabilities of radiologists and physicians, underscoring its clinical relevance in optimizing TACE planning. Reliance solely on pre-treatment CT often results in partial observation and suboptimal targeting due to heterogeneous pathological conditions. By incorporating MeWM’s recommended protocol, clinical decision-making is markedly enhanced, yielding performance improvements of  12.70, 11.45, 14.22, and 10.77 in F1-score, Jaccard, Precision, and Recall on our dataset. MeWM facilitates accurate tumor localization and enables predictive assessment of post-embolization outcomes, thereby reducing procedural uncertainty. 
Moreover, its synthetic post-treatment CT projections help anticipate embolization efficacy, optimize TACE distribution, and mitigate non-target embolization risks, contributing to enhanced therapeutic precision and individualized strategies. 
MeWM serves as a critical decision-support tool in interventional oncology, bridging anatomical imaging with functional assessment for meaningful clinical outcomes. 
As shown in Figure~\ref{fig:exp_tace}, interventional physicians refine TACE protocols by MeWM intervention, aligning treatment with expert practices.

\begin{table}[t]
  \centering
  \renewcommand{\arraystretch}{1.15}
  \caption{\textbf{Ablation studies of TACE protocol exploration on both datasets}. ``AM'' denotes Assistant Model, while ``CCL'' denotes Combo Contrastive Learning.
  Two components significantly contribute to better exploring the optimal treatment.}
  \vspace{-3mm}
  \resizebox{\linewidth}{!}{%
    \begin{tabular}{lcc|cc} 
    \toprule
    \multirow{2}{*}{\textbf{Metrics} $\uparrow$} & \multicolumn{2}{c|}{\textbf{In-house dataset}} & \multicolumn{2}{c}{\textbf{Public dataset}} \\ 
    & w/o AM & w/o CCL & w/o AM & w/o CCL  \\ 
    \hline
    F1-score  & 49.13 \textcolor{citegreen}{(-3.9)} & 50.97 \textcolor{citegreen}{(-1.4)} & 60.03 \textcolor{citegreen}{(-4.1)} & 62.90 \textcolor{citegreen}{(-1.2)}  \\
    Jaccard   & 35.40 \textcolor{citegreen}{(-3.2)} & 36.76 \textcolor{citegreen}{(-1.8)} & 45.36 \textcolor{citegreen}{(-3.1)} & 46.97 \textcolor{citegreen}{(-1.5)}  \\
    Precision & 56.39 \textcolor{citegreen}{(-6.7)} & 60.57 \textcolor{citegreen}{(-2.5)} & 75.00 \textcolor{blue}{(+2.4)} & 70.10 \textcolor{citegreen}{(-2.5)}  \\
    Recall    & 45.22 \textcolor{citegreen}{(-1.0)} & 45.36 \textcolor{citegreen}{(-0.8)} & 52.38 \textcolor{citegreen}{(-5.6)} & 56.72 \textcolor{citegreen}{(-2.2)}  \\
    \bottomrule
    \end{tabular}
  }
  \label{tab:ablation}
  % \vspace{-4mm}
\end{table}

\noindent\textbf{Ablation Study. }
As shown in Table~\ref{tab:ablation}, we ablate the effectiveness of Assitant Model and Combo Contrastive Learning (CCL) in TACE Protocol Exploration.
The results demonstrate that both the Assistant Model and Combo Contrastive Learning (CCL) contribute significantly to its performance. 
Removing the assistant model, which provides the location information of tumors for heuristic function, leads to a critical drop in F1-score (52.38$\rightarrow$49.13) on our dataset, as well as on public dataset. 
Similarly, omitting CCL reduces F1-score (\eg, 52.38$\rightarrow$50.97), indicating that CCL enhances the model's discrimination on action units. 
Overall, MeWM achieves the best results across all metrics, even outperforming radiologists in some areas, validating its effectiveness.

% \vspace{-3mm}
\section{Conclusion}
\label{sec:conclusion}
We present Medical World Model, which marks a step toward AI-driven precision medicine by simulating disease evolution and optimizing clinical strategies. 
By bridging generative modeling with medical decision-making, it enables a deeper understanding of treatment outcomes and refines intervention planning. 
The advancements of MeWM in tumor synthesis and survival analysis set the stage for future AI systems that seamlessly integrate with clinical workflows, driving the next generation of longitudinal data-driven healthcare.

% \clearpage
{\small
\bibliographystyle{ieee_fullname}
\bibliography{egbib}

\begin{thebibliography}{10}\itemsep=-1pt

\bibitem{deedsBCV}
deedsbcv.

\bibitem{anthropic2024claude}
AI Anthropic.
\newblock Claude 3.5 sonnet model card addendum.
\newblock {\em Claude-3.5 Model Card}, 2024.

\bibitem{Qwen2.5-VL}
Shuai Bai, Keqin Chen, Xuejing Liu, Jialin Wang, Wenbin Ge, Sibo Song, Kai Dang, Peng Wang, Shijie Wang, Jun Tang, Humen Zhong, Yuanzhi Zhu, Mingkun Yang, Zhaohai Li, Jianqiang Wan, Pengfei Wang, Wei Ding, Zheren Fu, Yiheng Xu, Jiabo Ye, Xi Zhang, Tianbao Xie, Zesen Cheng, Hang Zhang, Zhibo Yang, Haiyang Xu, and Junyang Lin.
\newblock Qwen2.5-vl technical report.
\newblock {\em arXiv preprint arXiv:2502.13923}, 2025.

\bibitem{bar2025navigation}
Amir Bar, Gaoyue Zhou, Danny Tran, Trevor Darrell, and Yann LeCun.
\newblock Navigation world models.
\newblock In {\em CVPR}, 2025.

\bibitem{bilic2023liver}
Patrick Bilic, Patrick Christ, Hongwei~Bran Li, Eugene Vorontsov, Avi Ben-Cohen, Georgios Kaissis, Adi Szeskin, Colin Jacobs, Gabriel Efrain~Humpire Mamani, Gabriel Chartrand, et~al.
\newblock The liver tumor segmentation benchmark (lits).
\newblock {\em Medical image analysis}, 84:102680, 2023.

\bibitem{billot2023synthseg}
Benjamin Billot, Douglas~N Greve, Oula Puonti, Axel Thielscher, Koen Van~Leemput, Bruce Fischl, Adrian~V Dalca, Juan~Eugenio Iglesias, et~al.
\newblock Synthseg: Segmentation of brain mri scans of any contrast and resolution without retraining.
\newblock {\em Medical image analysis}, 86:102789, 2023.

\bibitem{blattmann2023stable}
Andreas Blattmann, Tim Dockhorn, Sumith Kulal, Daniel Mendelevitch, Maciej Kilian, Dominik Lorenz, Yam Levi, Zion English, Vikram Voleti, Adam Letts, et~al.
\newblock Stable video diffusion: Scaling latent video diffusion models to large datasets.
\newblock {\em arXiv preprint arXiv:2311.15127}, 2023.

\bibitem{bluethgen2024vision}
Christian Bluethgen, Pierre Chambon, Jean-Benoit Delbrouck, Rogier van~der Sluijs, Ma{\l}gorzata Po{\l}acin, Juan~Manuel Zambrano~Chaves, Tanishq~Mathew Abraham, Shivanshu Purohit, Curtis~P Langlotz, and Akshay~S Chaudhari.
\newblock A vision--language foundation model for the generation of realistic chest x-ray images.
\newblock {\em Nature Biomedical Engineering}, pages 1--13, 2024.

\bibitem{bommasani2021opportunities}
Rishi Bommasani, Drew~A Hudson, Ehsan Adeli, Russ Altman, Simran Arora, Sydney von Arx, Michael~S Bernstein, Jeannette Bohg, Antoine Bosselut, Emma Brunskill, et~al.
\newblock On the opportunities and risks of foundation models.
\newblock {\em arXiv preprint arXiv:2108.07258}, 2021.

\bibitem{bruce2024genie}
Jake Bruce, Michael~D Dennis, Ashley Edwards, Jack Parker-Holder, Yuge Shi, Edward Hughes, Matthew Lai, Aditi Mavalankar, Richie Steigerwald, Chris Apps, et~al.
\newblock Genie: Generative interactive environments.
\newblock In {\em Forty-first International Conference on Machine Learning}, 2024.

\bibitem{busch2025multilingual}
Felix Busch, Philipp Prucker, Alexander Komenda, Sebastian Ziegelmayer, Marcus~R Makowski, Keno~K Bressem, and Lisa~C Adams.
\newblock Multilingual feasibility of gpt-4o for automated voice-to-text ct and mri report transcription.
\newblock {\em European Journal of Radiology}, 182:111827, 2025.

\bibitem{cardoso2022monai}
M~Jorge Cardoso, Wenqi Li, Richard Brown, Nic Ma, Eric Kerfoot, Yiheng Wang, Benjamin Murrey, Andriy Myronenko, Can Zhao, Dong Yang, et~al.
\newblock Monai: An open-source framework for deep learning in healthcare.
\newblock {\em arXiv preprint arXiv:2211.02701}, 2022.

\bibitem{chen2024huatuogpt}
Junying Chen, Chi Gui, Ruyi Ouyang, Anningzhe Gao, Shunian Chen, Guiming~Hardy Chen, Xidong Wang, Ruifei Zhang, Zhenyang Cai, Ke Ji, et~al.
\newblock Huatuogpt-vision, towards injecting medical visual knowledge into multimodal llms at scale.
\newblock {\em arXiv preprint arXiv:2406.19280}, 2024.

\bibitem{chen2024towards}
Qi Chen, Xiaoxi Chen, Haorui Song, Zhiwei Xiong, Alan Yuille, Chen Wei, and Zongwei Zhou.
\newblock Towards generalizable tumor synthesis.
\newblock In {\em Proceedings of the IEEE/CVF conference on computer vision and pattern recognition}, pages 11147--11158, 2024.

\bibitem{chen2024surgsora}
Tong Chen, Shuya Yang, Junyi Wang, Long Bai, Hongliang Ren, and Luping Zhou.
\newblock Surgsora: Decoupled rgbd-flow diffusion model for controllable surgical video generation.
\newblock {\em arXiv preprint arXiv:2412.14018}, 2024.

\bibitem{chen2023meditron}
Zeming Chen, Alejandro~Hern{\'a}ndez Cano, Angelika Romanou, Antoine Bonnet, Kyle Matoba, Francesco Salvi, Matteo Pagliardini, Simin Fan, Andreas K{\"o}pf, Amirkeivan Mohtashami, et~al.
\newblock Meditron-70b: Scaling medical pretraining for large language models.
\newblock {\em arXiv preprint arXiv:2311.16079}, 2023.

\bibitem{colosimo2002empirical}
Enrico Colosimo, Fla{\'{}}~vio Ferreira, Maristela Oliveira, and Cleide Sousa.
\newblock Empirical comparisons between kaplan-meier and nelson-aalen survival function estimators.
\newblock {\em Journal of Statistical Computation and Simulation}, 72(4):299--308, 2002.

\bibitem{costa2017end}
Pedro Costa, Adrian Galdran, Maria~Ines Meyer, Meindert Niemeijer, Michael Abr{\`a}moff, Ana~Maria Mendon{\c{c}}a, and Aur{\'e}lio Campilho.
\newblock End-to-end adversarial retinal image synthesis.
\newblock {\em IEEE transactions on medical imaging}, 37(3):781--791, 2017.

\bibitem{medgpt_page}
Michael D~Moor.
\newblock Medgpt, 2025.
\newblock Accessed: March 7, 2025.

\bibitem{du2023video}
Yilun Du, Mengjiao Yang, Pete Florence, Fei Xia, Ayzaan Wahid, Brian Ichter, Pierre Sermanet, Tianhe Yu, Pieter Abbeel, Joshua~B Tenenbaum, et~al.
\newblock Video language planning.
\newblock {\em arXiv preprint arXiv:2310.10625}, 2023.

\bibitem{du2024video}
Yilun Du, Mengjiao Yang, Pete Florence, Fei Xia, Ayzaan Wahid, Brian Ichter, Pierre Sermanet, Tianhe Yu, Pieter Abbeel, Joshua~B Tenenbaum, et~al.
\newblock Video language planning.
\newblock In {\em ICML}, 2024.

\bibitem{esser2021taming}
Patrick Esser, Robin Rombach, and Bjorn Ommer.
\newblock Taming transformers for high-resolution image synthesis.
\newblock In {\em Proceedings of the IEEE/CVF conference on computer vision and pattern recognition}, pages 12873--12883, 2021.

\bibitem{fox2002cox}
John Fox and Sanford Weisberg.
\newblock Cox proportional-hazards regression for survival data.
\newblock {\em An R and S-PLUS companion to applied regression}, 2002, 2002.

\bibitem{girdhar2023emu}
Rohit Girdhar, Mannat Singh, Andrew Brown, Quentin Duval, Samaneh Azadi, Sai~Saketh Rambhatla, Akbar Shah, Xi Yin, Devi Parikh, and Ishan Misra.
\newblock Emu video: Factorizing text-to-video generation by explicit image conditioning.
\newblock {\em arXiv preprint arXiv:2311.10709}, 2023.

\bibitem{guo2025deepseek}
Daya Guo, Dejian Yang, Haowei Zhang, Junxiao Song, Ruoyu Zhang, Runxin Xu, Qihao Zhu, Shirong Ma, Peiyi Wang, Xiao Bi, et~al.
\newblock Deepseek-r1: Incentivizing reasoning capability in llms via reinforcement learning.
\newblock {\em arXiv preprint arXiv:2501.12948}, 2025.

\bibitem{ha2018world}
David Ha and J{\"u}rgen Schmidhuber.
\newblock World models.
\newblock {\em arXiv preprint arXiv:1803.10122}, 2018.

\bibitem{hafner2023mastering}
Danijar Hafner, Jurgis Pasukonis, Jimmy Ba, and Timothy Lillicrap.
\newblock Mastering diverse domains through world models.
\newblock {\em arXiv preprint arXiv:2301.04104}, 2023.

\bibitem{hagag2024deep}
Amr Hagag, Ahmed Gomaa, Dominik Kornek, Andreas Maier, Rainer Fietkau, Christoph Bert, Yixing Huang, and Florian Putz.
\newblock Deep learning for cancer prognosis prediction using portrait photos by stylegan embedding.
\newblock In {\em International Conference on Medical Image Computing and Computer-Assisted Intervention}, pages 198--208. Springer, 2024.

\bibitem{hager2024evaluation}
Paul Hager, Friederike Jungmann, Robbie Holland, Kunal Bhagat, Inga Hubrecht, Manuel Knauer, Jakob Vielhauer, Marcus Makowski, Rickmer Braren, Georgios Kaissis, et~al.
\newblock Evaluation and mitigation of the limitations of large language models in clinical decision-making.
\newblock {\em Nature medicine}, 30(9):2613--2622, 2024.

\bibitem{hamamci2024ct2rep}
Ibrahim~Ethem Hamamci, Sezgin Er, and Bjoern Menze.
\newblock Ct2rep: Automated radiology report generation for 3d medical imaging.
\newblock In {\em International Conference on Medical Image Computing and Computer-Assisted Intervention}, pages 476--486. Springer, 2024.

\bibitem{hinrichs2016parametric}
Jan~B Hinrichs, Hoen-Oh Shin, Daniel Kaercher, Davut Hasdemir, Tim Murray, Till Kaireit, Carolin Lutat, Arndt Vogel, Bernhard~C Meyer, Frank~K Wacker, et~al.
\newblock Parametric response mapping of contrast-enhanced biphasic ct for evaluating tumour viability of hepatocellular carcinoma after tace.
\newblock {\em European radiology}, 26:3447--3455, 2016.

\bibitem{ho2020denoising}
Jonathan Ho, Ajay Jain, and Pieter Abbeel.
\newblock Denoising diffusion probabilistic models.
\newblock {\em Advances in Neural Information Processing Systems}, 33:6840--6851, 2020.

\bibitem{hu2023gaia}
Anthony Hu, Lloyd Russell, Hudson Yeo, Zak Murez, George Fedoseev, Alex Kendall, Jamie Shotton, and Gianluca Corrado.
\newblock Gaia-1: A generative world model for autonomous driving.
\newblock {\em arXiv preprint arXiv:2309.17080}, 2023.

\bibitem{hu2023label}
Qixin Hu, Yixiong Chen, Junfei Xiao, Shuwen Sun, Jieneng Chen, Alan~L Yuille, and Zongwei Zhou.
\newblock Label-free liver tumor segmentation.
\newblock In {\em Proceedings of the IEEE/CVF Conference on Computer Vision and Pattern Recognition}, pages 7422--7432, 2023.

\bibitem{huang2022common}
Pu Huang, Dengwang Li, Zhicheng Jiao, Dongming Wei, Bing Cao, Zhanhao Mo, Qian Wang, Han Zhang, and Dinggang Shen.
\newblock Common feature learning for brain tumor mri synthesis by context-aware generative adversarial network.
\newblock {\em Medical Image Analysis}, 79:102472, 2022.

\bibitem{huang2025ct}
Xiaoyu Huang, Yong Huang, Ping Li, and Kai Xu.
\newblock Ct-based deep learning predicts prognosis in esophageal squamous cell cancer patients receiving immunotherapy combined with chemotherapy.
\newblock {\em Academic Radiology}, 2025.

\bibitem{hurst2024gpt}
Aaron Hurst, Adam Lerer, Adam~P Goucher, Adam Perelman, Aditya Ramesh, Aidan Clark, AJ Ostrow, Akila Welihinda, Alan Hayes, Alec Radford, et~al.
\newblock Gpt-4o system card.
\newblock {\em arXiv preprint arXiv:2410.21276}, 2024.

\bibitem{isensee2021nnu}
Fabian Isensee, Paul~F Jaeger, Simon~AA Kohl, Jens Petersen, and Klaus~H Maier-Hein.
\newblock nnu-net: a self-configuring method for deep learning-based biomedical image segmentation.
\newblock {\em Nature methods}, 18(2):203--211, 2021.

\bibitem{katzman2018deepsurv}
Jared~L Katzman, Uri Shaham, Alexander Cloninger, Jonathan Bates, Tingting Jiang, and Yuval Kluger.
\newblock Deepsurv: personalized treatment recommender system using a cox proportional hazards deep neural network.
\newblock {\em BMC medical research methodology}, 18:1--12, 2018.

\bibitem{kim2024mdagents}
Yubin Kim, Chanwoo Park, Hyewon Jeong, Yik~Siu Chan, Xuhai Xu, Daniel McDuff, Hyeonhoon Lee, Marzyeh Ghassemi, Cynthia Breazeal, Hae Park, et~al.
\newblock Mdagents: An adaptive collaboration of llms for medical decision-making.
\newblock {\em Advances in Neural Information Processing Systems}, 37:79410--79452, 2024.

\bibitem{kirillov2023segment}
Alexander Kirillov, Eric Mintun, Nikhila Ravi, Hanzi Mao, Chloe Rolland, Laura Gustafson, Tete Xiao, Spencer Whitehead, Alexander~C Berg, Wan-Yen Lo, et~al.
\newblock Segment anything.
\newblock In {\em Proceedings of the IEEE/CVF international conference on computer vision}, pages 4015--4026, 2023.

\bibitem{lai2024pixel}
Yuxiang Lai, Xiaoxi Chen, Angtian Wang, Alan Yuille, and Zongwei Zhou.
\newblock From pixel to cancer: Cellular automata in computed tomography.
\newblock In {\em International Conference on Medical Image Computing and Computer-Assisted Intervention}, pages 36--46. Springer, 2024.

\bibitem{lecun2022path}
Yann LeCun.
\newblock A path towards autonomous machine intelligence version 0.9. 2, 2022-06-27.
\newblock {\em Open Review}, 62(1):1--62, 2022.

\bibitem{lee2024deep}
Kyung~Hwa Lee, Jungwook Lee, Gwang~Hyeon Choi, Jihye Yun, Jiseon Kang, Jonggi Choi, Kang~Mo Kim, and Namkug Kim.
\newblock Deep learning-based prediction of post-treatment survival in hepatocellular carcinoma patients using pre-treatment ct images and clinical data.
\newblock {\em Journal of Imaging Informatics in Medicine}, pages 1--12, 2024.

\bibitem{li2024endora}
Chenxin Li, Hengyu Liu, Yifan Liu, Brandon~Y Feng, Wuyang Li, Xinyu Liu, Zhen Chen, Jing Shao, and Yixuan Yuan.
\newblock Endora: Video generation models as endoscopy simulators.
\newblock In {\em International Conference on Medical Image Computing and Computer-Assisted Intervention}, pages 230--240. Springer, 2024.

\bibitem{li2025large}
Jia Li, Zichun Zhou, Han Lyu, and Zhenchang Wang.
\newblock Large language models-powered clinical decision support: enhancing or replacing human expertise?, 2025.

\bibitem{li2024text}
Xinran Li, Yi Shuai, Chen Liu, Qi Chen, Qilong Wu, Pengfei Guo, Dong Yang, Can Zhao, Pedro~RAS Bassi, Daguang Xu, et~al.
\newblock Text-driven tumor synthesis.
\newblock {\em arXiv preprint arXiv:2412.18589}, 2024.

\bibitem{liang2023deep}
Junhao Liang, Weisheng Zhang, Jianghui Yang, Meilong Wu, Qionghai Dai, Hongfang Yin, Ying Xiao, and Lingjie Kong.
\newblock Deep learning supported discovery of biomarkers for clinical prognosis of liver cancer.
\newblock {\em Nature Machine Intelligence}, 5(4):408--420, 2023.

\bibitem{liu2023clip}
Jie Liu, Yixiao Zhang, Jie-Neng Chen, Junfei Xiao, Yongyi Lu, Bennett A~Landman, Yixuan Yuan, Alan Yuille, Yucheng Tang, and Zongwei Zhou.
\newblock Clip-driven universal model for organ segmentation and tumor detection.
\newblock In {\em Proceedings of the IEEE/CVF International Conference on Computer Vision}, pages 21152--21164, 2023.

\bibitem{liu2024universal}
Jie Liu, Yixiao Zhang, Kang Wang, Mehmet~Can Yavuz, Xiaoxi Chen, Yixuan Yuan, Haoliang Li, Yang Yang, Alan Yuille, Yucheng Tang, et~al.
\newblock Universal and extensible language-vision models for organ segmentation and tumor detection from abdominal computed tomography.
\newblock {\em Medical image analysis}, 97:103226, 2024.

\bibitem{lu2024generative}
Taiming Lu, Tianmin Shu, Alan Yuille, Daniel Khashabi, and Jieneng Chen.
\newblock Generative world explorer.
\newblock {\em arXiv preprint arXiv:2411.11844}, 2024.

\bibitem{luo2024grounding}
Yunhao Luo and Yilun Du.
\newblock Grounding video models to actions through goal conditioned exploration.
\newblock {\em arXiv preprint arXiv:2411.07223}, 2024.

\bibitem{lyu2022pseudo}
Fei Lyu, Mang Ye, Jonathan~Frederik Carlsen, Kenny Erleben, Sune Darkner, and Pong~C Yuen.
\newblock Pseudo-label guided image synthesis for semi-supervised covid-19 pneumonia infection segmentation.
\newblock {\em IEEE Transactions on Medical Imaging}, 42(3):797--809, 2022.

\bibitem{moor2023foundation}
Michael Moor, Oishi Banerjee, Zahra Shakeri~Hossein Abad, Harlan~M Krumholz, Jure Leskovec, Eric~J Topol, and Pranav Rajpurkar.
\newblock Foundation models for generalist medical artificial intelligence.
\newblock {\em Nature}, 616(7956):259--265, 2023.

\bibitem{morshid2019machine}
Ali Morshid, Khaled~M Elsayes, Ahmed~M Khalaf, Mohab~M Elmohr, Justin Yu, Ahmed~O Kaseb, Manal Hassan, Armeen Mahvash, Zhihui Wang, John~D Hazle, et~al.
\newblock A machine learning model to predict hepatocellular carcinoma response to transcatheter arterial chemoembolization.
\newblock {\em Radiology: Artificial Intelligence}, 1(5):e180021, 2019.

\bibitem{nichol2021improved}
Alexander~Quinn Nichol and Prafulla Dhariwal.
\newblock Improved denoising diffusion probabilistic models.
\newblock In {\em International Conference on Machine Learning}, pages 8162--8171. PMLR, 2021.

\bibitem{radford2021learning}
Alec Radford, Jong~Wook Kim, Chris Hallacy, Aditya Ramesh, Gabriel Goh, Sandhini Agarwal, Girish Sastry, Amanda Askell, Pamela Mishkin, Jack Clark, et~al.
\newblock Learning transferable visual models from natural language supervision.
\newblock In {\em International conference on machine learning}, pages 8748--8763. PmLR, 2021.

\bibitem{rombach2022high}
Robin Rombach, Andreas Blattmann, Dominik Lorenz, Patrick Esser, and Bj{\"o}rn Ommer.
\newblock High-resolution image synthesis with latent diffusion models.
\newblock In {\em Proceedings of the IEEE/CVF conference on computer vision and pattern recognition}, pages 10684--10695, 2022.

\bibitem{seo2023masked}
Younggyo Seo, Danijar Hafner, Hao Liu, Fangchen Liu, Stephen James, Kimin Lee, and Pieter Abbeel.
\newblock Masked world models for visual control.
\newblock In {\em Conference on Robot Learning}, pages 1332--1344. PMLR, 2023.

\bibitem{shao2021transmil}
Zhuchen Shao, Hao Bian, Yang Chen, Yifeng Wang, Jian Zhang, Xiangyang Ji, et~al.
\newblock Transmil: Transformer based correlated multiple instance learning for whole slide image classification.
\newblock {\em Advances in neural information processing systems}, 34:2136--2147, 2021.

\bibitem{singhal2023large}
Karan Singhal, Shekoofeh Azizi, Tao Tu, S~Sara Mahdavi, Jason Wei, Hyung~Won Chung, Nathan Scales, Ajay Tanwani, Heather Cole-Lewis, Stephen Pfohl, et~al.
\newblock Large language models encode clinical knowledge.
\newblock {\em Nature}, 620(7972):172--180, 2023.

\bibitem{singhal2025toward}
Karan Singhal, Tao Tu, Juraj Gottweis, Rory Sayres, Ellery Wulczyn, Mohamed Amin, Le Hou, Kevin Clark, Stephen~R Pfohl, Heather Cole-Lewis, et~al.
\newblock Toward expert-level medical question answering with large language models.
\newblock {\em Nature Medicine}, pages 1--8, 2025.

\bibitem{sohl2015deep}
Jascha Sohl-Dickstein, Eric Weiss, Niru Maheswaranathan, and Surya Ganguli.
\newblock Deep unsupervised learning using nonequilibrium thermodynamics.
\newblock In {\em International conference on machine learning}, pages 2256--2265. PMLR, 2015.

\bibitem{team2023gemini}
Gemini Team, Rohan Anil, Sebastian Borgeaud, Jean-Baptiste Alayrac, Jiahui Yu, Radu Soricut, Johan Schalkwyk, Andrew~M Dai, Anja Hauth, Katie Millican, et~al.
\newblock Gemini: a family of highly capable multimodal models.
\newblock {\em arXiv preprint arXiv:2312.11805}, 2023.

\bibitem{tran2018closer}
Du Tran, Heng Wang, Lorenzo Torresani, Jamie Ray, Yann LeCun, and Manohar Paluri.
\newblock A closer look at spatiotemporal convolutions for action recognition.
\newblock In {\em Proceedings of the IEEE conference on Computer Vision and Pattern Recognition}, pages 6450--6459, 2018.

\bibitem{tulyakov2018mocogan}
Sergey Tulyakov, Ming-Yu Liu, Xiaodong Yang, and Jan Kautz.
\newblock Mocogan: Decomposing motion and content for video generation.
\newblock In {\em Proceedings of the IEEE conference on computer vision and pattern recognition}, pages 1526--1535, 2018.

\bibitem{radio_page}
Joost van Griethuysen~et. al.
\newblock Radiomics-based cox model, 2025.
\newblock Accessed: March 7, 2025.

\bibitem{wang2023towards}
X Wang, Z Zhu, G Huang, X Chen, and J~Drivedreamer Lu.
\newblock Towards real-world-driven world models for autonomous driving.
\newblock {\em arXiv preprint arXiv:2309.09777}, 2023.

\bibitem{wang2024drivedreamer}
Xiaofeng Wang, Zheng Zhu, Guan Huang, Xinze Chen, Jiagang Zhu, and Jiwen Lu.
\newblock Drivedreamer: Towards real-world-drive world models for autonomous driving.
\newblock In {\em European Conference on Computer Vision}, pages 55--72, 2024.

\bibitem{wu2023daydreamer}
Philipp Wu, Alejandro Escontrela, Danijar Hafner, Pieter Abbeel, and Ken Goldberg.
\newblock Daydreamer: World models for physical robot learning.
\newblock In {\em Conference on robot learning}, pages 2226--2240. PMLR, 2023.

\bibitem{xing2024cross}
Zhaohu Xing, Sicheng Yang, Sixiang Chen, Tian Ye, Yijun Yang, Jing Qin, and Lei Zhu.
\newblock Cross-conditioned diffusion model for medical image to image translation.
\newblock In {\em International Conference on Medical Image Computing and Computer-Assisted Intervention}, pages 201--211. Springer, 2024.

\bibitem{yao2021deepprognosis}
Jiawen Yao, Yu Shi, Kai Cao, Le Lu, Jianping Lu, Qike Song, Gang Jin, Jing Xiao, Yang Hou, and Ling Zhang.
\newblock Deepprognosis: Preoperative prediction of pancreatic cancer survival and surgical margin via comprehensive understanding of dynamic contrast-enhanced ct imaging and tumor-vascular contact parsing.
\newblock {\em Medical image analysis}, 73:102150, 2021.

\bibitem{yao2020deepprognosis}
Jiawen Yao, Yu Shi, Le Lu, Jing Xiao, and Ling Zhang.
\newblock Deepprognosis: Preoperative prediction of pancreatic cancer survival and surgical margin via contrast-enhanced ct imaging.
\newblock In {\em International Conference on Medical Image Computing and Computer-Assisted Intervention}, pages 272--282. Springer, 2020.

\bibitem{yao2021label}
Qingsong Yao, Li Xiao, Peihang Liu, and S~Kevin Zhou.
\newblock Label-free segmentation of covid-19 lesions in lung ct.
\newblock {\em IEEE transactions on medical imaging}, 40(10):2808--2819, 2021.

\bibitem{yao2025addressing}
Wenfang Yao, Chen Liu, Kejing Yin, William Cheung, and Jing Qin.
\newblock Addressing asynchronicity in clinical multimodal fusion via individualized chest x-ray generation.
\newblock {\em Advances in Neural Information Processing Systems}, 37:29001--29028, 2025.

\bibitem{zhang2023huatuogpt}
Hongbo Zhang, Junying Chen, Feng Jiang, Fei Yu, Zhihong Chen, Jianquan Li, Guiming Chen, Xiangbo Wu, Zhiyi Zhang, Qingying Xiao, et~al.
\newblock Huatuogpt, towards taming language model to be a doctor.
\newblock {\em arXiv preprint arXiv:2305.15075}, 2023.

\bibitem{zheng2024large}
Yanxin Zheng, Wensheng Gan, Zefeng Chen, Zhenlian Qi, Qian Liang, and Philip~S Yu.
\newblock Large language models for medicine: a survey.
\newblock {\em International Journal of Machine Learning and Cybernetics}, pages 1--26, 2024.

\bibitem{zhou2019high}
Yi Zhou, Xiaodong He, Shanshan Cui, Fan Zhu, Li Liu, and Ling Shao.
\newblock High-resolution diabetic retinopathy image synthesis manipulated by grading and lesions.
\newblock In {\em International conference on medical image computing and computer-assisted intervention}, pages 505--513. Springer, 2019.

\end{thebibliography}
}
% \clearpage

\clearpage
\appendix
\setcounter{page}{1}
% \maketitlesupplementary
\onecolumn
\renewcommand \thepart{}
\renewcommand \partname{}
\part{Appendix} % Start the appendix part
\setcounter{secnumdepth}{4}
\setcounter{tocdepth}{4}
\parttoc % Insert the appendix TOC

\clearpage

\section{Data Preprocessing}
\subsection{HCC-TACE-Seg dataset preprocessing}
The HCC-TACE-Seg dataset~\cite{morshid2019machine} refers to a single-institution collection of patients with confirmed hepatocellular carcinoma (HCC) who were treated at The University of Texas MD Anderson Cancer Center.
Data preprocessing for HCC-TACE-Seg involves resampling the provided CT images to a standardized spatial resolution while preserving the integrity of the original data structure. Specifically, images and masks are resampled to a target spacing of 0.8mm × 0.8mm × 3.0mm to standardize voxel dimensions across different cases.

\textbf{Longitudinal Registration}: Accurate image registration is essential to ensure that tumor boundaries are clearly defined across both the liver and HCC regions in different imaging modalities, such as arterial phase (AP) and portal venous phase (PVP) scans. The longitudinal registration process involves aligning the post-AP image to the pre-AP image, and the post-PVP to the pre-PVP image, addressing any misalignments between scans. Both linear and non-linear registration methods are employed through the open-sourced registration framework deedsBCV~\cite{deedsBCV} for optimal alignment.

\begin{figure*}[h]
\centering
\includegraphics[width=1.0\textwidth]{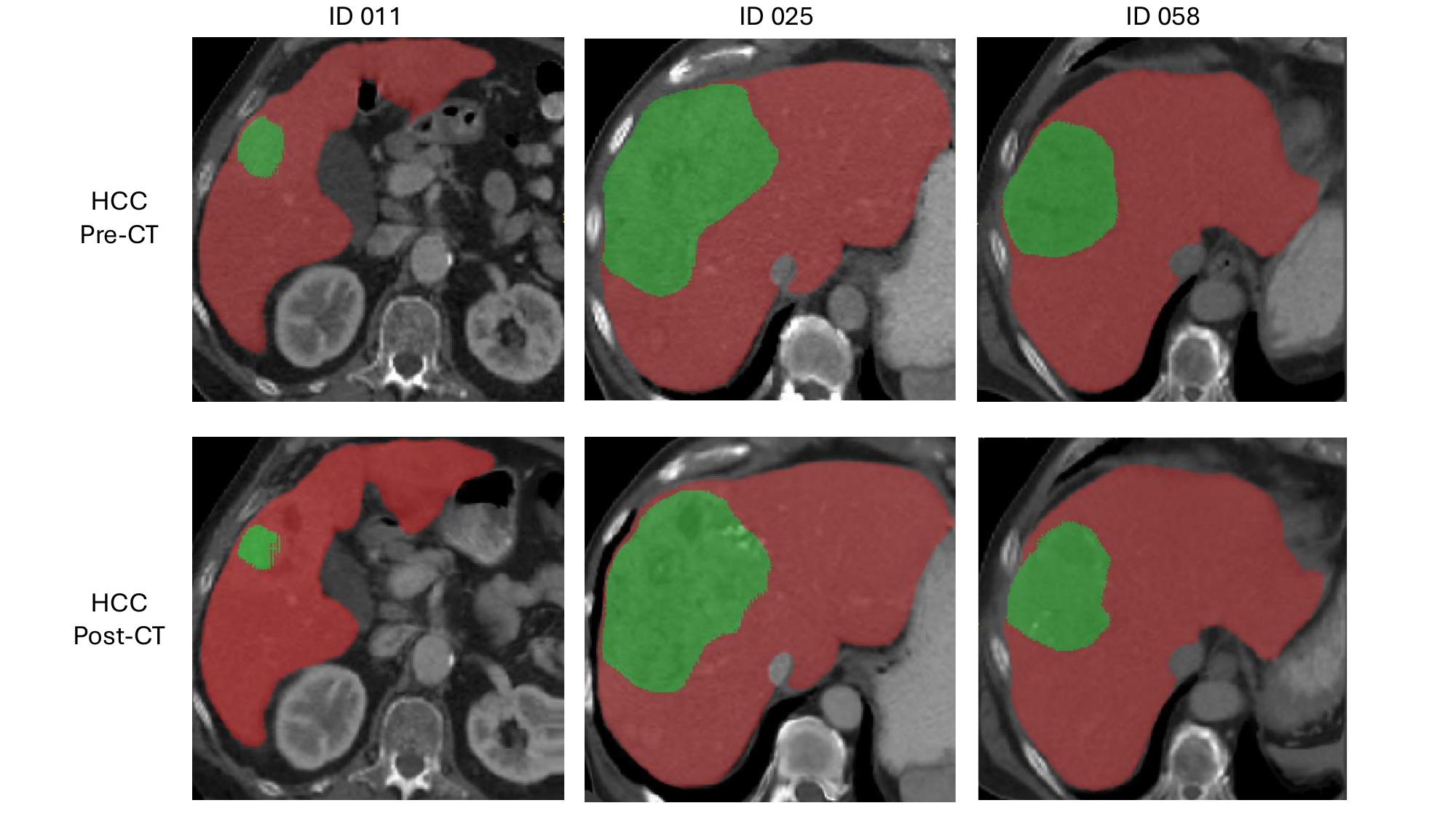}
% \vspace{-4mm}
\caption{Example of HCC-TACE-Seg dataset. The first row shows HCC Pre-CT images, and the second row shows HCC Post-CT images. The red mask represents the liver, while the green mask represents the HCC tumor.
}
\label{fig:HCC-TACE-Seg images}
\vspace{-4mm}
\end{figure*}

\textbf{Liver and HCC Cancer Segmentation}: We utilize a nnUNet-based~\cite{isensee2021nnu} mode trained on the public LiTS dataset~\cite{bilic2023liver} for liver and HCC cancer segmentation. For postprocessing, we adopt connected component analysis to extract the liver and HCC regions precisely. This approach ensures that the tumor and liver boundaries are defined clearly, which is crucial for downstream analysis. An example of pre- and post-treatment CT images, along with liver and tumor segmentation generated from the HCC-TACE-Seg dataset, is shown in~\Cref{fig:HCC-TACE-Seg images}.

We also conduct a Component Size Filtering strategy. 
A component size filtering step is applied, with a minimum threshold of 300 voxels, ensuring the accurate identification of tumor and liver regions. This step helps to remove noise or irrelevant small regions, improving the precision of segmentation results.
Only the image data paired with the following meta-information are selected for further analysis:
\begin{itemize}
    \item Chemotherapy: Information about whether the patient underwent chemotherapy treatment, including details about the type of chemotherapy regimen.
    \item Overall survival: Overall survival time in months.
    \item Survival status: 0 indicates that the patient is alive or lost to follow-up, while 1 indicates death. Details are in~\Cref{tab:HCC-TACE-Seg_data}.
\end{itemize}

\begin{table*}[h]
\renewcommand{\arraystretch}{1.5}
    \centering
    \begin{tabular}{ccccc}
        \toprule
        \textbf{Patient ID} & \textbf{Chemotherapy} & \textbf{Overall Survival (months)} & \textbf{Survival Status} \\
        \midrule
        HCC\_009 & Cisplatin; Doxorubicin; Mitomycin; Lipiodol  & 4.7 & 1.0 \\
        HCC\_011 & Cisplatin; Doxorubicin; Mitomycin; Lipiodol  & 19.3 & 1.0 \\
        HCC\_025 & Cisplatin; Doxorubicin; Mitomycin; Lipiodol  & 30.0 & 1.0 \\
        HCC\_034 & Doxorubicin; Lipiodol; LC beads & 18.9 & 1.0 \\
        HCC\_042 & Cisplatin; Mitomycin; Lipiodol  & 34.1 & 1.0 \\
        HCC\_051 & Cisplatin; Mitomycin; Lipiodol  & 12.9 & 1.0 \\
        HCC\_058 & Cisplatin; Mitomycin; Lipiodol  & 87.0 & 0.0 \\
        HCC\_067 & Cisplatin; Doxorubicin; Mitomycin; Lipiodol  & 90.9 & 0.0 \\
        HCC\_079 & Doxorubicin; LC beads; Lipiodol  & 42.5 & 1.0 \\
        HCC\_091 & Doxorubicin; LC beads; Lipiodol  & 25.3 & 0.0 \\
       
        \bottomrule
    \end{tabular}
    \caption{An example of HCC-TACE-Seg dataset metadata, including chemotherapy, overall survival, and survival status. For survival status, a value of 0 indicates that the patient is alive or lost to follow-up, while a value of 1 indicates death.}
    \label{tab:HCC-TACE-Seg_data}
\end{table*}

\begin{table*}[]
\renewcommand{\arraystretch}{1.5}
\centering
\setlength{\tabcolsep}{8pt} % Adjust column spacing
\begin{tabular}{c p{8cm} c c}
    \toprule
    \textbf{Patient ID} & \centering \textbf{Processed Chemotherapy} & \textbf{OS (months)} & \textbf{Survival Status} \\
    \midrule
    HCC\_08116730 & Raltitrexed 4 mg was infused through the catheter; 5 ml ultra-liquid Lipiodol and 5 mg Epirubicin were mixed to create an emulsion for embolization; the emulsion was slowly injected under fluoroscopic guidance; an appropriate amount of Gelatin Sponge particles was used to embolize the tumor-feeding branches of the S8 segment of the right hepatic artery; Lipiodol deposition in the tumor was satisfactory; tumor-feeding arteries were occluded on the final angiography. & 1.4 & 0.0 \\\hline
    HCC\_01061677 & THP 10 mg was infused through the catheter; 10 mg THP and 10 ml ultra-liquid Lipiodol were mixed to create an emulsion for embolization; 12 ml of the emulsion was slowly injected under fluoroscopic guidance; Lipiodol deposition in the tumor and satellite lesions was satisfactory; tumor-feeding arteries were occluded on the final angiography.  & 75.7 & 1.0 \\\hline
    HCC\_01192613 & THP 40 mg and 30 ml ultra-liquid Lipiodol, along with a small amount of contrast agent, were mixed to create an emulsion for embolization; 30 ml of the emulsion was slowly injected under fluoroscopic guidance; a small amount of Gelatin Sponge particles was used for embolization; Lipiodol deposition in the tumor was satisfactory; no tumor staining was observed on the final angiography. & 84.6 & 1.0 \\\hline
    HCC\_01204059 & Cisplatin 40 mg was infused through the catheter; 10 ml ultra-liquid Lipiodol was slowly injected under fluoroscopic guidance for embolization of the right hepatic artery tumor-feeding branches; 3 ml ultra-liquid Lipiodol was injected for protective embolization of the segment II branch of the left hepatic artery; Lipiodol deposition in the tumor was acceptable; tumor staining mostly disappeared on the final angiography.  & 17.1 & 0.0 \\\hline
    HCC\_01532843 & Oxaliplatin 100 mg and Epirubicin 30 mg were infused through the catheter; 10 mg Epirubicin and 10 ml ultra-liquid Lipiodol were mixed to create an emulsion for embolization; 10 ml of the emulsion was slowly injected under fluoroscopic guidance; an appropriate amount of Gelatin Sponge particles was used to embolize the tumor-feeding branches of the right hepatic artery; Lipiodol deposition in the tumor was satisfactory; tumor staining disappeared on the final angiography. & 29.3 & 1.0 \\\hline
      HCC\_01532843 & Epirubicin 40 mg and Oxaliplatin 100 mg were infused through the catheter; 10 ml ultra-liquid Lipiodol was slowly injected under fluoroscopic guidance for embolization; Lipiodol deposition in the tumor was satisfactory; tumor-feeding arteries were mostly occluded on the final angiography. & 21.6 & 1.0 \\
    
    \bottomrule
\end{tabular}
    \caption{An example of HCC-TACE dataset metadata, including chemotherapy, overall survival, and survival status. For survival status, a value of 0 indicates that the patient is alive or lost to follow-up, while a value of 1 indicates death.}
    \label{tab:HCC-TACE_data}
\end{table*}

\begin{figure*}[t]
\centering
\includegraphics[width=1.0\textwidth]{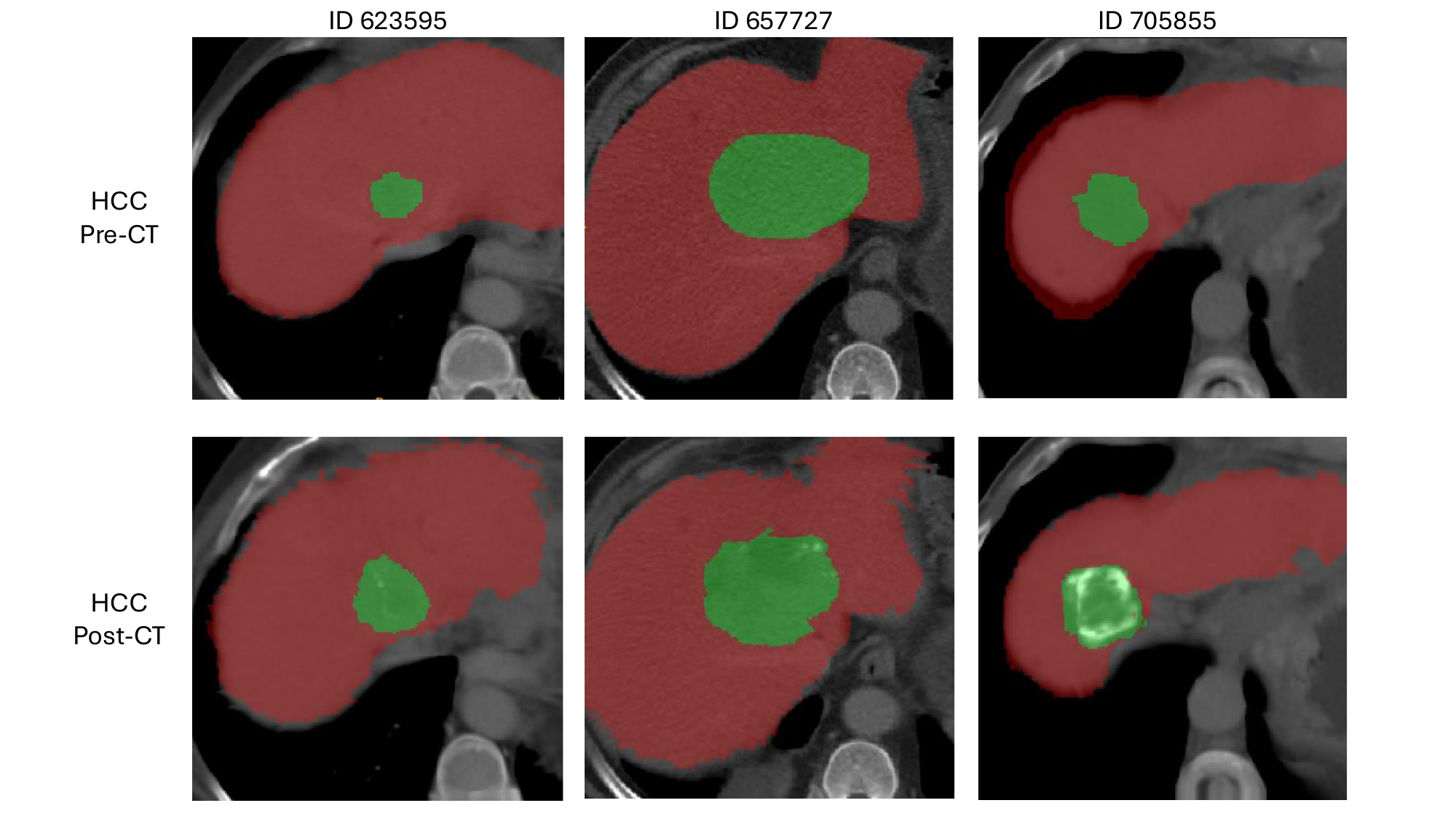}
% \vspace{-4mm}
\caption{Example of HCC-TACE dataset. The first row shows HCC Pre-CT images, and the second row shows HCC Post-CT images. The red mask represents the liver, while the green mask represents the HCC tumor. In post-treatment CT imaging of HCC, particularly after Transarterial Chemoembolization (TACE), the viable tumor region and its enhancement intensity decrease due to Lipiodol accumulation and treatment-induced necrosis. Lipiodol appears hyperdense (bright) on post-treatment CT, indicating areas that have been successfully embolized.}
\label{fig:HCC-TACE images}
\vspace{-4mm}
\end{figure*}

\subsection{HCC-TACE dataset preprocessing}

The HCC-TACE dataset is a large-scale, self-collected repository containing 338 longitudinal pairs of pre- and post-treatment CT scans, along with well-annotated liver and tumor masks, as well as clinical records. These records include TACE radiotherapy reports (considered the gold action) and Overall Survival (OS) time. Details are presented in~\Cref{tab:HCC-TACE_data}. The dataset is split into training (including validation) and testing sets in a 9:1 ratio. All images and masks are resampled to a target spacing of 0.8mm × 0.8mm × 3.0mm to standardize voxel dimensions across different cases.

\textbf{Longitudinal Registration}: We also employ deedsBCV~\cite{deedsBCV} to align the post-AP image with the pre-AP image, and the post-PVP image with the pre-PVP image, addressing any misalignments between the scans.

\textbf{Liver and HCC Cancer Annotation}:  In this dataset, all liver and tumor masks for each CT scan are carefully annotated by radiologists. For postprocessing, we also apply connected component analysis to accurately extract the liver and HCC regions. An example of pre- and post-treatment CT images, along with liver and tumor segmentation generated from the HCC-TACE dataset, is shown in~\Cref{fig:HCC-TACE images}.

% \newpage

\section{Implementation Details}

% \noindent\textbf{Policy Model: }
\subsection{Policy Model}
\begin{figure*}[t]
\centering
\includegraphics[width=0.7\textwidth]{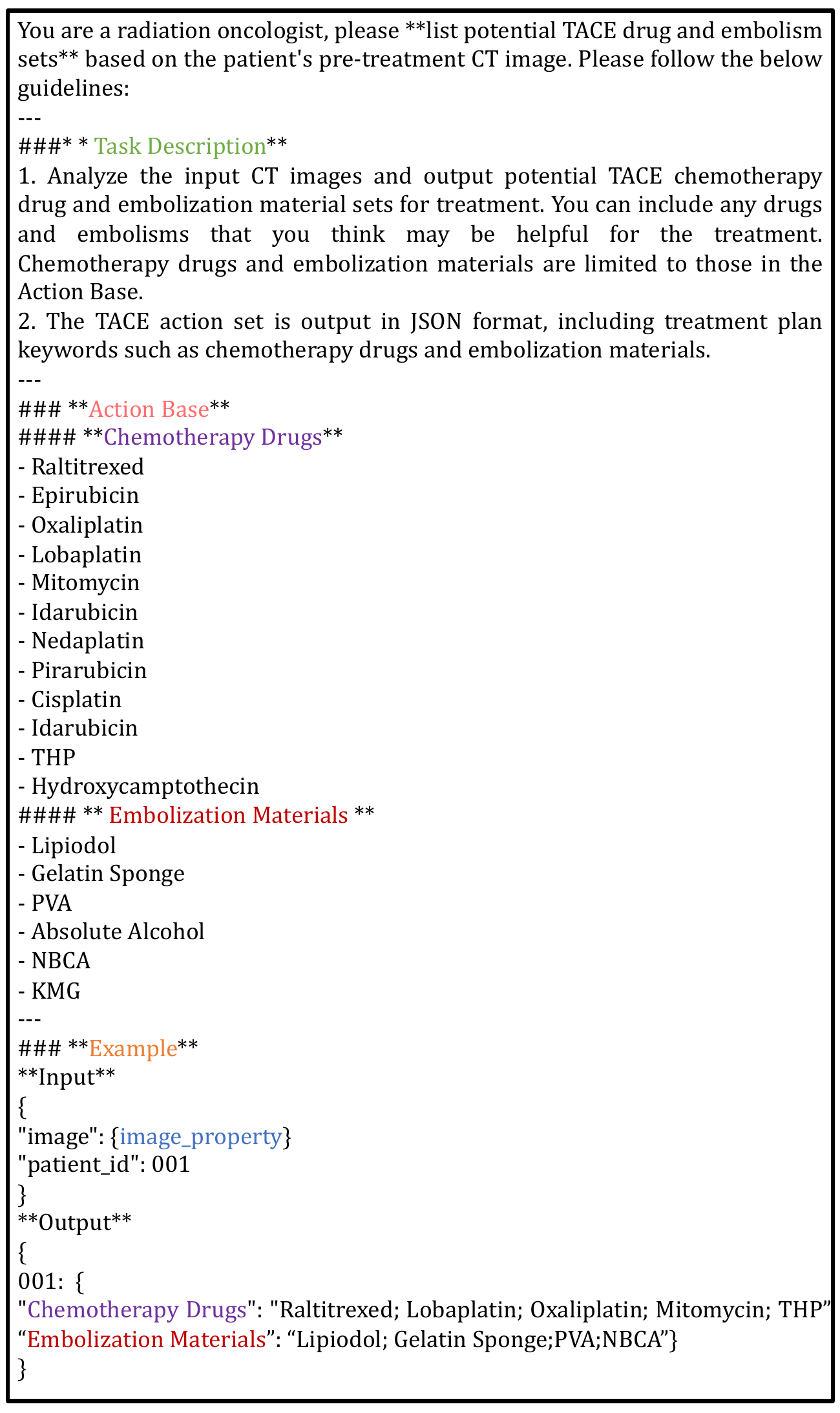}
% \vspace{-4mm}
\caption{Policy model prompt template for our dataset.
}
\label{fig:policy1}
\vspace{-4mm}
\end{figure*}

We adopt GPT-4o to obtain the initial observation from the given pre-treatment CT scans and collect the individualized potential drugs and embolism during TACE treatment. 
An example is presented in Figure~\ref{fig:policy1}.
Then, we refine the action set using DeepSeek-R1~\cite{guo2025deepseek}, which reasons the clinical conflicts in the current action set and summarizes a better action set using clinical guidelines for individuals (\eg, Multiple platinum-based drugs cannot be used simultaneously).

% \noindent\textbf{Dynamics Model: }
\subsection{Dynamics Model}
In this study, we implement Dynamics Model by training the corresponding Diffusion Model~\cite{li2024text} specifically from pre-treatment liver tumors to post-treatment liver tumors. 
The CT scans are oriented according to specific axcodes and resampled to achieve isotropic spacing of $1.0 \times 1.0 \times 1.0~\text{mm}^3$. 
$96 \times 96 \times 96$ patches are randomly cropped around either foreground voxels based on a set ratio.
Their intensities are truncated to the range [$-175, 600$] to maintain the discrimination of lipiodol/necrosis/viable areas~\cite{hinrichs2016parametric}, then linearly normalized to [-1, 1].
We utilize the Adam optimizer with hyperparameters $\beta_1=0.9$ and $\beta_2=0.999$, a learning rate of 0.0001, and a batch size of 10 per GPU. 
The training is conducted on A6000 GPUs for 2 days, over a total of 2,000 iterations.

% \smallskip\noindent\textbf{Assistant Model: }
\subsection{Assistant Model}
We employ a nnUNet-based~\cite{isensee2021nnu}  segmentation model for the segmentation of liver and tumor in post-treatment CT.
As suggested by Chen~\etal~\cite{chen2024towards}, we generate realistic tumor-like shapes using ellipsoids, and combine these generated tumor masks with the healthy CT volumes to create a range of realistic liver tumors.
We pre-train the model on the generated and real tumors for robust generalization. 
Then, we finetune it on post-treatment CT scans as well as liver and tumor masks. The implementation is in Python, leveraging MONAI\footnote{Cardoso~\etal~\cite{cardoso2022monai}: \href{https://monai.io/}{https://monai.io/}}. The CT scans are oriented according to specific axcodes and resampled to achieve isotropic spacing of $1.0 \times 1.0 \times 1.0~\text{mm}^3$. Their intensities are truncated to the range [$-175, 600$] to maintain the discrimination of lipiodol/necrosis/viable areas, then linearly normalized to [-1, 1].
During training, $96 \times 96 \times 96$ patches are randomly cropped around either foreground or background voxels based on a set ratio. Each patch is subjected to a $90^\circ$ rotation with probability 0.1 and an intensity shift of 0.1 with probability 0.2. To avoid confusing the left and right organs, mirroring augmentation is not used.

The model is initialized with pre-trained liver tumor weights from DiffTumor~\cite{chen2024towards}, then fine-tuned on our dataset for 2,000 epochs. We set the base learning rate to 0.0002 and use a batch size of 8, along with a linear warmup and a cosine annealing schedule. Training spans 2 days on eight A6000 GPUs. Additional details on the tumor synthesis process during Segmentation Model training can be found in DiffTumor~\cite{chen2024towards}.

For inference, a sliding window strategy with 0.75 overlap is used. Tumor predictions that fall outside their corresponding organs are removed by post-processing with organ pseudo-labels obtained from previous research\footnote{Liu~\etal~\cite{liu2023clip,liu2024universal}: \href{https://github.com/ljwztc/CLIP-Driven-Universal-Model}{https://github.com/ljwztc/CLIP-Driven-Universal-Model}}.

% \noindent\textbf{Heuristic Function: }
\subsection{Heuristic Function}
We implement a CNN-based survival analysis model as Heuristic Function.
The framework adopts 3D ResNet (MC3) as the backbone and Two-way Transformer as the interaction module of pre-treatment and post-treatment CT features. A multi-instance aggregator~\cite{shao2021transmil} with consecutive fully connected layers is utilized for survival risk scoring.
It is implemented on PyTorch using 8 NVIDIA RTX A6000 GPUs.
Their intensities are truncated to the range [$-175, 600$] to maintain the discrimination of lipiodol/necrosis/viable areas, then linearly normalized to [-1, 1].
During training, $96 \times 96 \times 96$ patches are randomly cropped around either foreground or background voxels based on a set ratio. Each patch is subjected to a $90^\circ$ rotation with probability 0.1 and an intensity shift of 0.1 with probability 0.2. To avoid confusing the left and right organs, mirroring augmentation is not used.
We utilize the Adam optimizer with hyperparameters $\beta_1=0.9$ and $\beta_2=0.999$, a learning rate of 0.00002, and a batch size of 5 per GPU. 
For the inference of each case, we predict the survival risk scores of 5 patches around the foreground and average them to obtain the final score.

\section{Comparison against Multi-Modal GPTs}
\begin{figure*}[t]
\centering
\includegraphics[width=0.8\textwidth]{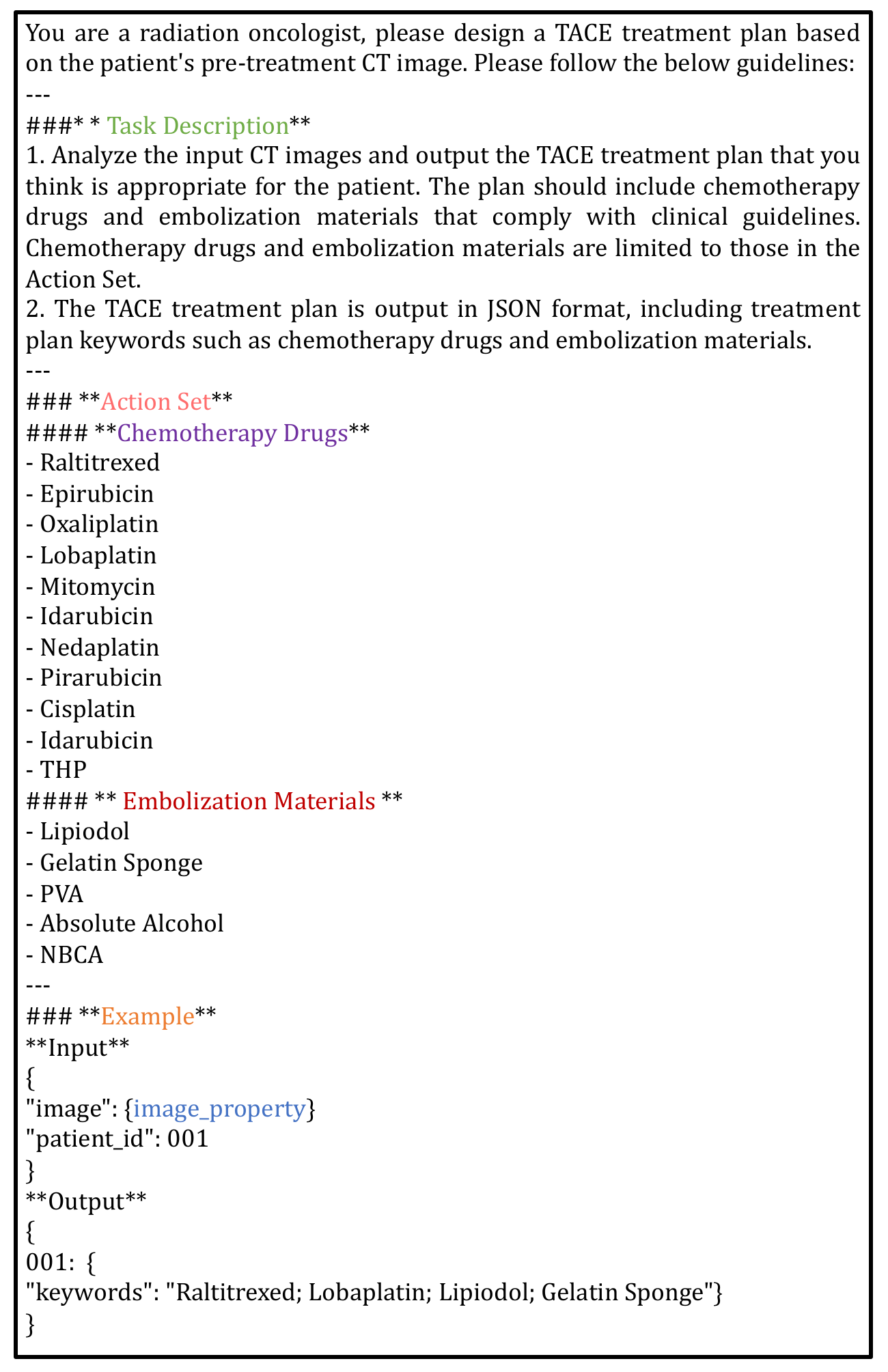}
% \vspace{-4mm}
\caption{VLM prompt template for our dataset.
}
\label{fig:prompt1}
\vspace{-4mm}
\end{figure*}

\begin{figure*}[t]
\centering
\includegraphics[width=0.8\textwidth]{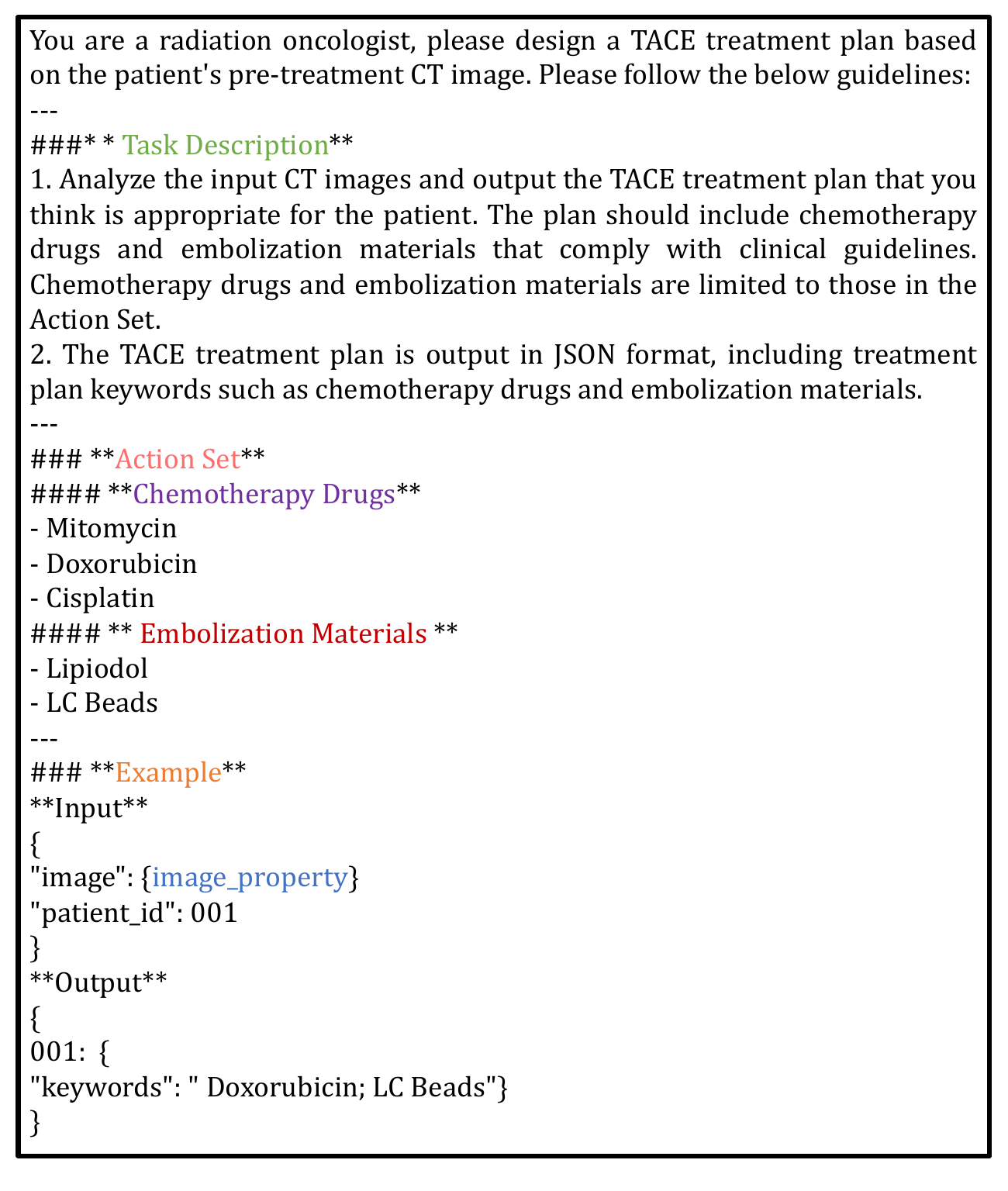}
% \vspace{-4mm}
\caption{VLM prompt template for HCC-TACE-Seg.
}
\label{fig:prompt2}
\vspace{-4mm}
\end{figure*}

We carefully design prompt templates for multi-modal GPTs to generate TACE treatment protocol. 
The first template (Figure~\ref{fig:prompt1}) is for our dataset with a larger action space, defining the task description, a predefined set of chemotherapy drugs and embolization materials, and an example of input-output in JSON format. 
The second template (Figure~\ref{fig:prompt2}) is specifically designed for the HCC-TACE-Seg dataset, featuring a different selection of chemotherapy drugs and embolization materials. 
Both prompts instruct GPTs to analyze CT images and generate an appropriate TACE treatment plan, submitting results in JSON format with predefined keywords.

\end{document}